\pdfoutput=1

\documentclass[11pt]{article}

\usepackage{acl}

\usepackage{times}
\usepackage{latexsym}

\usepackage[T1]{fontenc}

\usepackage[utf8]{inputenc}

\usepackage{microtype}

\usepackage{booktabs}
\usepackage{array}
\usepackage{geometry}
\usepackage{multirow}
\usepackage{graphicx}   
\usepackage{colortbl}   
\usepackage{xcolor}  
\usepackage{adjustbox}

\usepackage{multirow}
\usepackage{makecell}
\usepackage{array}
\usepackage{booktabs}
\usepackage{graphicx}
\usepackage{amssymb}
\usepackage{multirow}
\usepackage{rotating}
\usepackage{tabularx}
\usepackage{enumitem}

\usepackage{xcolor}
\usepackage{pifont} 
\newcommand{\cmark}{\ding{51}} 
\newcommand{\xmark}{\ding{55}} 

\newcommand{\Observation}[2]{%
  \noindent\textbf{Observation~#1}\nolinebreak[4]\ (\textit{#2})%
}

\newcommand{\good}{\textcolor{green!60!black}{\cmark}}
\newcommand{\bad}{\textcolor{red!70!black}{\xmark}}

\newcommand{\equal}{\textsuperscript{\dag}}   
\newcommand{\corr}{\textsuperscript{*}}       

\title{\textsc{VocalBench-DF}: A Benchmark for Evaluating Speech LLM Robustness to Disfluency}

\author{Hongcheng Liu\equal, Yixuan Hou\equal, Heyang Liu, Yuhao Wang, \\ \textbf{Yanfeng Wang, Yu Wang\corr} \\
Shanghai Jiao Tong University \\ \texttt{\{hongcheng\_liu,yixuan303a,liuheyang,colane \}@sjtu.edu.cn} \\ \texttt{\{wangyanfeng, yuwangsjtu \}@sjtu.edu.cn} \\}

\begin{document}
\maketitle
\begingroup
\renewcommand{\thefootnote}{\fnsymbol{footnote}}
\footnotetext[2]{Equal contributions.}
\footnotetext[1]{Corresponding authors.}
\endgroup
\begin{abstract}
While Speech Large Language Models (Speech-LLMs) show strong performance in many applications, their robustness is critically under-tested, especially to speech disfluency. Existing evaluations often rely on idealized inputs, overlooking common disfluencies, particularly those associated with conditions like Parkinson's disease. This work investigates \textit{\textbf{whether current Speech-LLMs can maintain performance when interacting with users who have speech impairments.}} To facilitate this inquiry, we introduce \textsc{VocalBench-DF}, a framework for the systematic evaluation of disfluency across a multi-dimensional taxonomy. Our evaluation of 22 mainstream Speech-LLMs reveals substantial performance degradation, indicating that their real-world readiness is limited. Further analysis identifies phoneme-level processing and long-context modeling as primary bottlenecks responsible for these failures. Strengthening recognition and reasoning capability from components and pipelines can substantially improve robustness.
These findings highlight the urgent need for new methods to improve disfluency handling and build truly inclusive Speech-LLMs

\end{abstract}
\section{Introduction}
Speech large language models (Speech-LLMs) have demonstrated strong capabilities in many domains, particularly in speech-driven interaction~\cite{arora2025landscape,chu2024qwen2audio}. By directly taking speech signals as input, Speech-LLMs can achieve lower latency and provide more fluent user experiences, thereby improving the efficiency and accessibility of interactive systems~\cite{wang2025vocalnet,li2025baichuan}. Moreover, Speech-LLMs are able to capture human intentions across diverse genders and speaking styles, making them valuable tools for users worldwide~\cite{ma2024leveraging}.

\begin{figure}
    \centering
    \includegraphics[width=0.8\linewidth]{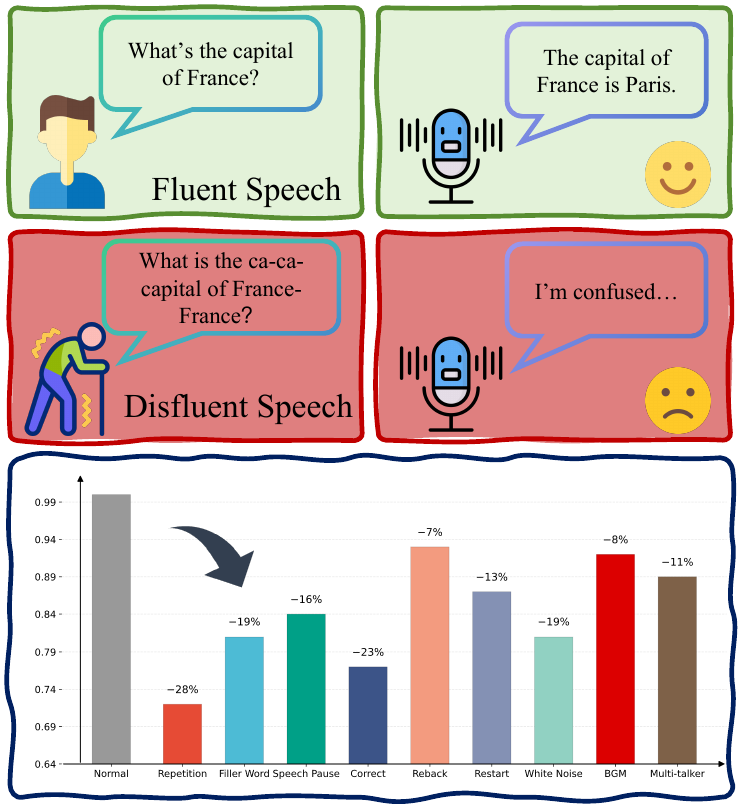}
    \caption{Illustration of the current speech-LLM's various performance under fluent and disfluent speech. The bottom part presents the GLM-4-Voice's performance under various conditions, where almost all disfluent types exhibit huge degradation.}
    \label{fig:case}
    \vspace{-2em}
\end{figure}

\begin{table*}[t]
\centering
\renewcommand{\arraystretch}{1} 
\small
\adjustbox{width=0.98\textwidth,center}{
\begin{tabular}{c c c c c c c}
\toprule
\multirow{2}{*}{\thead{Dataset}} 
 & \multicolumn{4}{c}{Linguistic Realization Disfluency} & \multicolumn{2}{c}{Interaction Interference Disfluency} \\
\cmidrule(lr){2-5} \cmidrule(lr){6-7} 
 & \thead{Phoneme} 
 & \thead{Word} 
 & \thead{Sentence} 
 & \thead{Prosodic} 
 & \thead{Natural Environment} 
 & \thead{Person Talking} \\
\toprule
VoiceBench~\cite{chen2024voicebench}  &  \bad       &     \good       &      \good     &       \bad &       \good     &  \bad       \\
MMSU~\cite{wang2025mmsu}& \bad & \good     &     \good       &  \bad   &  \bad   &    \bad           \\
VocalBench~\cite{liu2025vocalbench}  &  \bad       &  \bad&   \bad  &   \bad  &      \good      & \bad  \\
WildSpeech-Bench~\cite{zhang2025wildspeech}        &  \bad    &  \bad      &   \bad  &   \bad  &    \good  & \good  \\
\textsc{VocalBench-DF}~(ours)        & \good & \good & \good & \good & \good &   \good      \\
\bottomrule
\end{tabular}}%
\caption{Comparison of benchmarks across different types of disfluency. Others focus on interaction disfluency and little linguistic disfluency, but ignore the speech signal itself. Instead, ours establishes a systemic benchmark of speech fluency, especially including the speech-oriented tasks.}
\label{tab:disfluency}
\vspace{-1.5em}
\end{table*}

Nevertheless, real-world speech input is far from ideal. Beyond environmental noise and other external factors, a central challenge is fluency: many speakers worldwide experience disfluent speech due to medical conditions such as Parkinson’s disease or stroke~\cite{zheng2025interspeech2025speechaccessibility}. For example, stroke survivors and people with speech disorders often produce fragmented or atypical utterances. This raises an important, underexplored question: \textbf{Can current Speech-LLMs maintain consistent performance when interacting with users who have speech impairments?} Several benchmarks address related aspects, VocalBench targets noisy speech~\cite{liu2025vocalbench}, and VoiceBench focuses on sentence correction~\cite{chen2024voicebench}. However, as summarized in Table~\ref{tab:disfluency}, existing efforts remain fragmented and tend to overlook core speech-internal phenomena, particularly phonemic structure, prosodic patterns, and speech overlapping. 

To address this gap, we present \textsc{VocalBench-DF}, the first benchmark that systematically evaluates model performance across diverse speech disfluency types. We organize disfluency along two major dimensions: linguistic realization of speech and the process of interaction interference. These dimensions are further decomposed into six fine-grained subcategories and nine subtasks. We place particular emphasis on fluency properties of the speech signal, including phonemic structure, prosodic patterns, and speech overlapping. Taken together, these components provide a holistic and comprehensive framework for assessing disfluency in speech-based interaction.

Based on this, we conduct a comprehensive evaluation of 22 mainstream Speech-LLMs spanning diverse architectures and training strategies. We reveal that most state-of-the-art models are surprisingly brittle. We uncover fundamental limitations in handling disfluencies at the phoneme level, maintaining long-context dependencies, and recovering user intent during conversational interruptions. However, our analysis also identifies key pathways toward more robust models, highlighting the significant positive impact of stronger speech recognition and complex reasoning from both component models and training pipelines.
Our contributions are threefold.
\begin{itemize}[itemsep=0pt, topsep=0pt, parsep=0pt]
    \item \textbf{Systematic categorization.} We establish a new paradigm for systematic disfluency evaluation by introducing a multi-dimensional framework that categorizes phenomena along two axes: linguistic realization and interactional interference.
    \item \textbf{\textsc{VocalBench-DF} benchmark.} We develop and release \textsc{VocalBench-DF}, the first benchmark designed to rigorously assess Speech-LLMs across diverse disfluency types with a special emphasis on speech-signal characteristics (e.g., phoneme structure, prosody pattern, and speech overlapping).
    \item \textbf{Comprehensive analysis.} We evaluate 22 mainstream Speech-LLMs, revealing that most models are surprisingly brittle. We further uncover the fundamental limitations and key pathways toward more robustness.
\end{itemize}


\begin{figure*}[t]
    \centering
    \includegraphics[width=.95\linewidth]{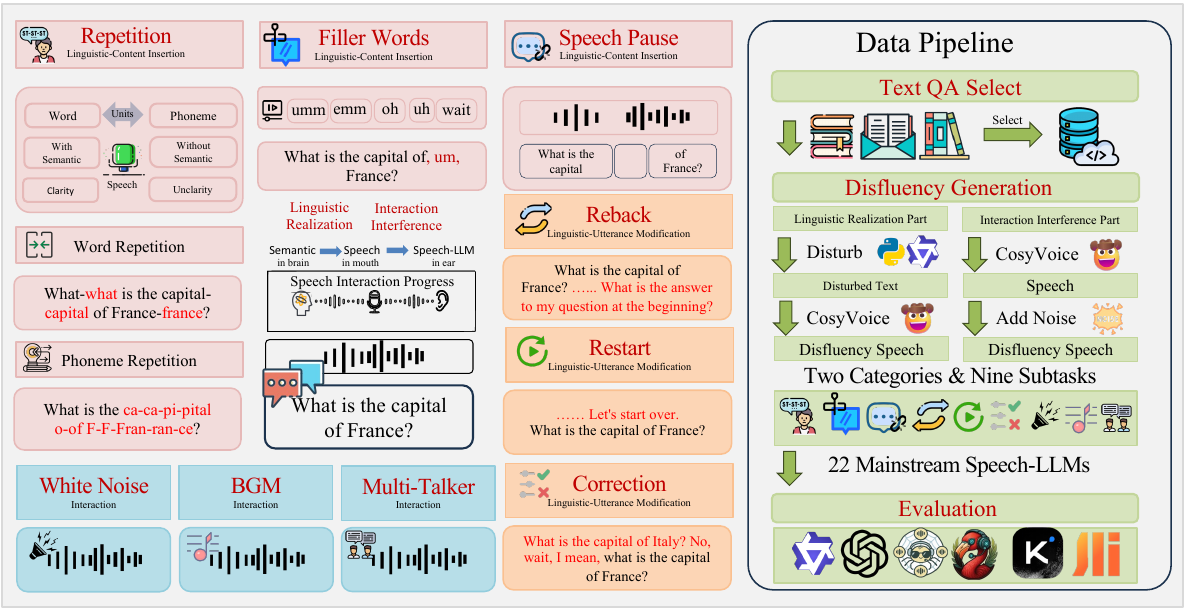}
    \caption{The demonstration of our dataset. The left part is the details about two categories and nine sub-tasks, and the right part is the whole pipeline of data construction and evaluation.}
    \label{fig:overall}
    \vspace{-1.5em}
\end{figure*}

\section{Related Works}

Many speech benchmarks have been proposed to assess the capabilities of Speech-LLMs. For instance, MMAR~\cite{ma2025mmar} targets deep reasoning, whereas VocalBench~\cite{liu2025vocalbench} emphasizes vocal performance. AudioBench~\cite{wang2024audiobench} and OpenAudioBench~\cite{li2025baichuan} extend evaluation to linguistic, paralinguistic, and multimodal understanding. More recently, URO-Bench~\cite{yan2025urobenchcomprehensiveevaluationendtoend} provides a comprehensive suite for end-to-end spoken dialogue in speech-to-speech settings, covering multilingualism and multi-round dialogue. However, these benchmarks center on linguistic competence and task success while giving limited attention to performance under real-world disfluency.

To address this gap, several benchmarks such as C3~\cite{ma2025c3} and VoiceBench~\cite{chen2024voicebench} evaluate models under disfluent or noisy input, as summarized in Table~\ref{tab:disfluency}. Yet, these efforts devote less attention to speech-internal phenomena, leaving open the question of whether users with inherently disfluent speech can interact effectively with Speech-LLMs. In this work, we introduce \textsc{VocalBench-DF}, a benchmark specifically designed to systematically evaluate Speech-LLMs under diverse disfluency conditions, achieved by isolating and controlling distinct speech-oriented phenomena, including phoneme structure, prosody pattern, and speech overlapping.

\section{Benchmark Construction}
According to the process of speech-based interaction, we categorize disfluency in two distinct but related dimensions: linguistic realization disfluency and interactional interference disfluency. Linguistic realization refers to the speaker’s own performance, where speech production is disrupted by phenomena such as repetition or frequent pauses. Interactional interference, by contrast, arises during the communication process and is often caused by environmental factors, such as background noise or overlapping speech, which reduce the smoothness and coherence of information exchange. The details can be found in Figure~\ref{fig:overall}. 
\subsection{Linguistic Realization Disfluency}
The realization of Language can be analyzed at different linguistic levels, ranging from local to global phenomena~\cite{skidmore-moore-2022-incremental}. Local disfluencies typically involve short insertions or disruptions within the flow of speech, such as repetitions and filled pauses. In contrast, global disfluencies affect the structure of the entire utterance, as seen in restarts and corrections. To illustrate this distinction, the following subsections take the sentence ``What is the capital of France?'' as the reference sentence.
\subsubsection{Content Insertion}
\paragraph{Repetition}
Repetition is a type of disfluency characterized by repeated units, which may occur at the level of phonemes or words, for example, ``What is the ca-ca-capital of France-France?''. This phenomenon can distort the intended meaning of words and sentences, especially when repetitions occur in key terms, and often creates difficulties for speech-LLMs. Its unpredictable occurrence and variable duration make it particularly difficult to handle and result in noticeable disfluency. In our experiments, we simulate repetition by randomly inserting repeated phonemes or words into the original sentence, and then generate the corresponding speech using text-to-speech models.

\paragraph{Filler Words}
Filler words are very common in everyday speech and are often used to fill silence or adjust the prosodic of language, such as `uh', `um', or `you know'. For example, ``What is the capital of, um, France?''. Although these words usually carry minimal semantic content and mainly serve to prevent conversational breakdowns, they objectively interrupt the flow of semantic expression and cause breaks in meaning, which may influence the output of speech-LLMs and lead to significant disfluency. In our experiments, we simulate the filler words using the same method as repetition.
\paragraph{Speech Pause}
Pauses are another common form of filler, frequently observed in daily conversations. Unlike filler words, a pause occurs as a silence between two words and does not introduce any additional lexical content into the sentence. However, an unexpected pause can still disrupt the rhythm of the utterance and break the coherence between the words before and after, thereby affecting the overall semantic flow and generating a substantial disfluency. In practice, we randomly insert silences between the two words in speech with various durations to simulate the pause phoneme.
\subsubsection{Utterance Modification}
\paragraph{Reback}
Reback refers to the phenomenon of abruptly returning to an original or earlier point in a conversation, for example, ``What is the capital of France? $\dots$ What is the answer to my question at the beginning?''. This does not involve modifying the original sentence itself, but the intervening utterance may introduce irrelevant information and cause substantial disfluency. In our experiments, we generated irrelevant sentences with Qwen3-8B and inserted them after the original question. 
\paragraph{Restart}
Restart is contrastive to the reback, which restarts the dialogue to the later sentence, for example, ``$\dots$ Let's start over. What is the capital of France?''.  Although the irrelevant information is placed before the question, such disfluency still increases the processing burden on the model and can easily cause semantic misalignment in the generated responses. In our experiments, we generate the additional content in the same way as reback, but place it before the original question.

\paragraph{Correction}
Correction is similar to the restart in that the dialogue is redirected to a later utterance, but the intervening content takes the form of another question rather than a declarative sentence, with a short context. For example: ``What is the capital of Italy? No, wait, I mean, what is the capital of France?''. In this case, the model is more likely to be misled into answering the earlier, incorrect question. In our experiments, we simulate corrections by randomly selecting another question and inserting it before the original one.

\subsection{Interaction Interference Disfluency}
Since interactional interference disfluency is mainly caused by environmental factors and overlapping speech, we select three representative interference factors for analysis.
\paragraph{White Noise}
White noise is one of the most common forms of noise in real-world environments, typically originating from sources such as electrical current or wind. This type of noise can significantly interfere with information transmission during speech interaction, resulting in obvious disfluency. In our implementation, we simulate this effect by adding normally distributed white noise to the speech vectors.
\paragraph{Background Music}
Background music (BGM) is another common source of environmental interference, often present in public spaces such as shopping malls or transportation hubs. Unlike white noise, background music contains structured rhythms and melodies, which may overlap with speech frequencies and mask important phonetic cues. This interference can distract speech-LLMs heavily, reducing the accuracy and fluency of speech interaction. In our implementation, we simulate this effect by mixing music tracks of varying styles into the original speech.
\paragraph{Multi-talker}
Multi-talker interference occurs when multiple speakers talk simultaneously, producing overlapping speech segments. This phenomenon is frequent in real conversations, especially in group discussions or crowded environments. Overlapping speech makes it difficult to separate the target utterance from irrelevant voices, thereby increasing recognition errors and causing interactional disfluency. In our experiments, we simulate multi-talker interference by mixing additional speech from other speakers in a lower voice.

\section{Experiments}

\begin{table}[t]
  \centering
  \renewcommand{\arraystretch}{1} 
  \resizebox{.49\textwidth}{!}{
     \begin{tabular}{lcc}
      \toprule
      \textbf{Model Name} & \textbf{Speech-Encoder}& \textbf{Based-LLM} \\ 
      \midrule
      GPT-4o-audio~\citeyearpar{gpt-audio}           & -- & -- \\
      Mini-omni-S~\citeyearpar{xie2024mini,mini2}     & Whisper-small & Qwen2-0.5B \\
      VocalNet-1B/8B~\citeyearpar{wang2025vocalnet}        & Whisper-large-v3  & LLaMA3-series\\
      Qwen2-Audio-7B~\citeyearpar{chu2024qwen2audio}        & Whisper-large-v3 & Qwen-7B \\
      LLaMA-Omni2-S~\citeyearpar{fang2025llama}    & Whisper-large-v3 & Qwen2.5-series  \\
      Qwen2.5-Omni~\citeyearpar{xu2025qwen25omni}      & Whisper-large-v3 & Qwen2.5-series \\
        VocalNet-qwen25-7B~\citeyearpar{wang2025vocalnet}        & Whisper-large-v3  & Qwen2.5-7B \\
      MiniCPM-o-2.6~\citeyearpar{yao2024minicpm}       & Whisper-medium-300M & Qwen2.5-7B \\
      Audio-Flamigo-3~\citeyearpar{goel2025flamingo3}     & AF-Whisper & Qwen2.5-7B \\ 
      Kimi-Audio~\citeyearpar{ding2025kimi}        & Kimi-Audio-Tokenizer & Qwen2.5-7B \\
      Baichuan-Omni-1.5~\citeyearpar{li2025baichuan}   & Baichuan-Audio-Tokenizer & Qwen2.5-7B \\
      GLM-4-Voice~\citeyearpar{zeng2024glm} & GLM-4-Voice-Tokenizer & GLM-4-9B\\
      Qwen3-Omni-S~\citeyearpar{Qwen3-Omni}   & AuT encoder & Qwen3-30B-A3B \\
      \bottomrule
    \end{tabular}
  }
  \caption{Comparison of Speech-LLMs. The -S denotes the -series for convenience. As there are other task-unrelated architectures, such as the talker and visual encoder, we only show the task-oriented component.}
  \label{tab:model}
  \vspace{-1.5 em}
\end{table}

\subsection{Experimental Settings}
In our experiments, we adopt a subset of OpenAudioBench~\footnote{\url{https://huggingface.co/datasets/baichuan-inc/OpenAudioBench}} as the source dataset, which includes LLaMA questions, trivia QA, and web questions, amounting to a total of 2300 samples. And each disfluency type is based on all the samples. For the speech generation process, we employ CosyVoice~\cite{du2024cosyvoice} as the speech synthesizer, taking textual questions as input and randomly selecting prompt voices, to enhance diversity and make it closer to the real world. To evaluate model performance, we use accuracy under the normal setting as the baseline metric, reflecting the original capability of models, and define the relative accuracy rate as an indicator of their ability to handle various disfluent input.

\subsection{Evaluated Models}
We conducted a comprehensive evaluation on 22 representative speech-LLMs, with the detailed specifications summarized in Table~\ref{tab:model}. These models vary considerably in terms of model size, architectural design, and training methodology, which allows us to systematically examine how each factor influences model behavior across different categories of disfluencies. Furthermore, we use 0.1 as the temperature rate and evaluate the models in RTX-3090 and NVIDIA A100. More details can be found in Appendix~\ref{app-sec:models}.

\subsection{Dataset Quality}
\begin{table}[t]
\centering
\resizebox{.4\textwidth}{!}{
\begin{tabular}{lccc}
\hline
\rowcolor{gray!20} 
\textbf{Model} & \textbf{Text} & \textbf{Wav2Vec} & \textbf{Whisper} \\
\hline
GPT-4o-mini & 60.83 & 54.65 & 57.00  \\
GPT-4o & 64.39 & 58.13  & 60.70 \\
\hline
\end{tabular}
}
\caption{Scores and relative errors for Text, Wav2Vec result, and Whisper result across different models.}
\label{tab:gpt-asr}
\vspace{-1.5em}
\end{table}

\begin{table*}[t]
\centering
\resizebox{\textwidth}{!}{
\begin{tabular}{lccccccccccc}
    \toprule
    \textbf{Model} & \textbf{Normal} & \makecell{\textbf{Repetition}} & \makecell{\textbf{Filler Word}} & \makecell{\textbf{Speech Pause}} & \textbf{Correct} & \textbf{Reback} & \textbf{Restart} & \makecell{\textbf{White Noise}} & \textbf{BGM} & \makecell{\textbf{Multi-talker}} & \textbf{Average} \\
    \midrule
    \rowcolor{gray!15} \multicolumn{12}{c}{\textbf{Human Score}} \\
    Human Score & 99.50 & 0.97 & 0.99 & 0.98 & 1.00 & 1.00 & 1.00 & 0.94 & 0.95 & 0.96 & 0.97
     \\ \hline
    \rowcolor{gray!15} \multicolumn{12}{c}{\textbf{ASR + LLM}} \\ 
    Whisper+GPT-4o & 60.70 & 0.83 & 0.96 & 0.94 & 0.99 & 1.01 & 0.99 & 0.94 & 0.98 & 0.94 & 0.95 \\
    \midrule
    \rowcolor{gray!15} \multicolumn{12}{c}{\textbf{LLM Scale} $\leq$ 3B} \\
    Mini-Omni & 11.26 & 0.60 & 0.70 & 0.81 & 0.19 & 0.60 & 0.23 & 0.84 & 0.96 & 0.82 & 0.64 \\
    Mini-Omni2 & 13.17 & 0.64 & 0.78 & 0.83 & 0.18 & 0.78 & 0.25 & 0.88 & 0.98 & 0.90 & 0.69 \\
    LLaMA-Omni2-0.5B & 14.87 & 0.63 & 0.83 & 0.89 & 0.41 & 0.93 & 0.25 & 0.93 & 0.99 & 0.77 & 0.74 \\
    VocalNet-1B & 29.83 & 0.64 & 0.86 & 0.88 & 0.58 & 0.75 & 0.22 & 0.93 & 0.97 & 0.84 & 0.74 \\
    LLaMA-Omni2-1.5B & 27.48 & 0.68 & 0.91 & 0.86 & 0.68 & 0.91 & 0.23 & 0.88 & 0.95 & 0.75 & 0.76 \\
    LLaMA-Omni2-3B & 31.39 & 0.79 & 0.94 & 0.89 & 0.74 & 0.99 & 0.20 & 0.90 & 0.95 & 0.81 & 0.80 \\
    Qwen2.5-Omni-3B & 35.22 & 0.75 & 0.92 & 0.93 & 0.86 & 0.98 & 0.90 & 0.96 & 0.98 & 0.95 & 0.91 \\
    \rowcolor{gray!15} \textbf{Average} & 
    \textbf{23.32} 	&\textbf{0.68 }	&\textbf{0.85 }	&\textbf{0.87} 	&\textbf{0.52} 	&\textbf{0.85} &	\textbf{0.33} 	&\textbf{0.90} 	&\textbf{0.97} 	&\textbf{0.83 }	&\textbf{0.75} \\
    \midrule
    \rowcolor{gray!15} \multicolumn{12}{c}{3B $<$ \textbf{LLM Scale} $\leq$ 8B} \\
    Qwen2-Audio-7B & 30.87 & 0.53 & 0.74 & 0.81 & 0.43 & 0.76 & 0.17 & 0.95 & 1.00 & 0.88 & 0.70 \\
    Qwen2.5-Omni-7B & 41.61 & 0.85 & 0.94 & 0.95 & 0.91 & 0.99 & 0.95 & 0.98 & 0.99 & 0.94 & 0.94 \\
    LLaMA-Omni2-7B & 36.96 & 0.76 & 0.91 & 0.86 & 0.83 & 0.95 & 0.19 & 0.87 & 0.98 & 0.82 & 0.80 \\
    VocalNet-qwen25-7B & 39.00 &0.80 &0.93 &0.90 &0.98 &0.97 &0.22 &0.92 &0.94& 0.86 &0.84\\
    MiniCPM-o-2.6 & 46.57 & 0.79 & 0.92 & 0.94 & 0.85 & 0.99 & 0.92 & 0.90 & 0.96 & 0.92 & 0.91 \\
    Audio-Flamigo-3 &36.09& 0.80 &0.87 &0.87 &0.68& 0.97& 0.90 &0.90 &0.97 &0.90&0.87\\
    Kimi-Audio & 45.52 & 0.87 & 0.97 & 0.99 & 0.99 & 1.05 & 1.03 & 0.96 & 0.99 & 0.98 & 0.98 \\
    Baichuan-Omni-1.5 & 41.57 & 0.75 & 0.89 & 0.89 & 0.75 & 0.97 & 0.90 & 0.91 & 0.98 & 0.92 & 0.88 \\
    VocalNet-8B & 43.52 & 0.69 & 0.86 & 0.86 & 0.88 & 0.69 & 0.21 & 0.94 & 0.97 & 0.84 & 0.77 \\
    \rowcolor{gray!15} \textbf{Average} & \textbf{40.19} &	\textbf{0.76} 	&\textbf{0.89} &	\textbf{0.90} 	&\textbf{0.81} &	\textbf{0.93} 	&\textbf{0.61} &	\textbf{0.93} &\textbf{0.98 }&	\textbf{0.90} &	\textbf{0.85} \\
    \midrule
    \rowcolor{gray!15} \multicolumn{12}{c}{\textbf{LLM Scale} $>$ 8B} \\
    GLM-4-Voice &38.35 &0.72& 0.81 &0.84& 0.77 &0.93 &0.87& 0.81 &0.92 &0.89&0.84 \\
    LLaMA-Omni2-14B & 44.26 & 0.80 & 0.93 & 0.85 & 0.88 & 0.96 & 0.19 & 0.91 & 0.97 & 0.81 & 0.81 \\
    Qwen3-Omni-Instruct &53.52& 0.85& 0.95& 0.96& 0.98& 0.99& 0.29& 0.95& 0.98& 0.94&0.88\\
    Qwen3-Omni-Captioner &51.17&0.87 &0.95 &0.98 &0.95 &0.80 &0.19 &0.97 &1.00 &0.94&0.85\\
    Qwen3-Omni-Thinking &60.65& 0.91& 0.97& 0.99& 0.89& 0.94& 0.28& 0.97& 0.99& 0.95&0.88 \\
    \rowcolor{gray!15} \textbf{Average} & \textbf{49.59}&
    \textbf{0.83}&
    \textbf{0.92}&
    \textbf{0.92}&
    \textbf{0.89}&
    \textbf{0.92}&
    \textbf{0.36}&
    \textbf{0.92}&
    \textbf{0.97}&
    \textbf{0.91}&
    \textbf{0.85}\\
    \midrule
    \rowcolor{gray!15} \multicolumn{12}{c}{\textbf{Speech-LLM} (closed-source)} \\
    GPT-4o-audio & 61.74 & 0.94 & 0.99 & 0.99 & 0.95 & 1.00 & 0.99 & 0.20 & 0.96 & 0.97 & 0.89 \\\midrule
    \rowcolor{gray!15} \textbf{Average} &
    \textbf{--} & \textbf{0.76} & \textbf{0.89} & \textbf{0.90} &
    \textbf{0.74} & \textbf{0.90} & \textbf{0.48} & \textbf{0.88} &
    \textbf{0.97} & \textbf{0.88} & \textbf{0.82} \\

    \bottomrule
\end{tabular}
}
\caption{Model performance on various disfluencies. ``Normal'' denotes by accuracy; others are relative to Normal. As there are other task-unrelated architectures in models, we split these models by LLM scales.}
\label{tab:main-result}
\vspace{-1.5em}
\end{table*}

\paragraph{Normal}
To assess normal-speech quality, we evaluate GPT-4o and GPT-4o-mini using both the original text inputs and the corresponding automatic speech recognition transcripts (ASR). As shown in Table~\ref{tab:gpt-asr}, the absolute score gap between text and ASR inputs is below \(3\) points, indicating high transcript fidelity with minimal recognition-induced degradation. Aggregating across tasks, we obtain an overall normal-speech quality of \(99.5\%\) by human and WER of \(4.77\%\) by Whisper-large-v3, further corroborating this conclusion.

\paragraph{Disfluency}
To ensure the reliability of the generated disfluent speech, we conduct a human evaluation of the audio queries. We randomly sample \(10\%\) of cases from each type and rate them as good~(\(1\) score), normal~(\(0.5\) score), or bad~(\(0\) score). Since our goal is to assess speech quality, annotators only judge whether the intended semantics are perceivable. As summarized in Table~\ref{tab:main-result}, all quality scores are higher than \(0.94\) with the average of \(0.97\), suggesting that these speech preserves semantic clarity for human listeners.

\section{Experimental Results}

The primary results on different types of disfluent input are presented in Table~\ref{tab:main-result}. Several key conclusions can be drawn from the results.

\subsection{Disfluencies Challenge Current Models}
\Observation{1}{Existing speech-LLMs lag behind human and ASR+LLM pipelines.}
Compared with human raters and ASR+LLM pipelines, Speech-LLMs show a substantial deficit, particularly for repetition and sentence-level reasoning. In our evaluation, Speech-LLMs obtain an average score of \(0.82\), far below the human score of \(0.97\), indicating persistent difficulty in detecting and repairing disfluencies. The gap also persists under utterance modifications: Whisper+GPT-4o maintains accuracy across all three modification types, with performance comparable to the normal setting, whereas Speech-LLMs degrade under the same conditions. These findings highlight open challenges in modeling real-world conversational dynamics, where disfluencies and on-the-fly revisions are common. 


\Observation{2}{Stable performance across disfluencies remains elusive.}
The results reveal substantial dispersion at the disfluency phenomenon levels.  The scores of the disfluency type range from \(0.97\) down to \(0.48\), with an average of \(0.82\). Crucially, most of the evaluated systems cannot maintain consistently high scores across all categories; models that excel at some phenomena often falter on others. This pattern points to differential sensitivity to specific cues rather than uniform noise, highlighting the necessity for integrating disfluency-aware mechanisms in model training to ensure more stable performance across disfluency categories.


\subsection{Comparative Task-Level Performance.}
\Observation{3}{Linguistic-based is more difficult than interaction-based disfluency.}
As shown in the table, disfluency arising from interactional interference has a limited impact: nearly all models score above \(0.88\) across the three conditions. In contrast, disfluencies on linguistic realization exert a much larger effect, with about half of the types averaging below \(0.76\). These results suggest that existing models maintain relatively robust performance under environmental influence but struggle to recover the original semantics in the presence of linguistic disfluency. This finding highlights the need to strengthen semantic understanding and disfluency-aware modeling for current models.

\Observation{4}{Models preserve meaning under non-semantic disfluency.}
Among linguistic realization disfluencies, those that perturb prosodic without altering lexical content, specifically filler words and speech pauses, have a relatively small impact, each averaging around \(0.90\). Importantly, these phenomena preserve the original words and do not introduce new semantics. This suggests that existing models can recover the intended meaning when the disfluency is semantically neutral.


\Observation{5}{Repetition is the most consistently harmful.}
Repetition emerges as the most uniformly adverse factor: nearly all open-source models score below \(0.85\), revealing a broad and systematic weakness. By duplicating whole words or phonemic segments, repetition perturbs meaning, distorting compositional cues and creating spurious associations. Unlike filler words, which largely preserve speakers' intent, repetitions more frequently alter the interpretation semantics, driving models away from the original proposition. These findings call for methods that reliably detect and down-weight repeated material while safeguarding the intended meaning.

\Observation{6}{Restart disfluency is strongly bimodal.}
Performance on restart disfluency exhibits a clear bimodality: failures cluster below \(0.25\), whereas successful cases exceed \(0.90\). Notably, models that underperform on restart, however, often handle other utterance-level modifications well. For example, VocalNet-qwen25-7B reaches \(0.98\) on query revisions (correct) and \(0.97\) on long-context (reback), yet only \(0.22\) on restart. A common failure mode is anchoring early to false starts rather than reconciling them with the final query. These results motivate targeted training strategies for contradiction-aware restart handling.



\subsection{Component and Training are important.}

\Observation{7}{Scaling alone is insufficient.}
Performance depends strongly on the choice of LLM, especially on the average scores. Within a single family, scores typically rise with scale under both clean and disfluent inputs; for example, within the LLaMA-Omni2 series, the 14B model substantially outperforms smaller variants across most categories, indicating that scaling helps under disfluency. However, the trend is not universal. In our VocalNet variants, VocalNet-8B uses a larger LLM than VocalNet-qwen25-7B with the same speech encoder and training pipeline, yet the larger model performs worse. This non-monotonic behavior suggests that parameter count alone is insufficient; the intrinsic capability of the LLM and its fitness are the key determinants of performance.

\Observation{8}{Stronger speech encoders yield better performance.}
Regarding the LLM backbone fixed at Qwen2.5-7B. Kimi-Audio uses Kimi-Audio-Tokenizer (Whisper-Large) and attains a score of \(0.98\) on average, but MiniCPM-o-2.6 uses Whisper-Medium and reaches \(0.91\). The same things happen on the Baichuan-Omni-1.5 (\(0.88\)) and LLaMA-Omni2-7B (\(0.80\)), indicating that a stronger speech encoder materially improves performance under disfluency. 

\Observation{9}{Training procedures are critical.}
Beyond component choice, the training pipeline substantially affects consistency under disfluency. Holding architecture fixed, VocalNet-qwen25-7B and Qwen2.5-Omni-7B share the same LLM backbone and the same speech encoder; their overall scores differ markedly: \(0.84\) and \(0.94\). The latter’s larger training datasets and three-stage alignment, from speech-encoder training to LLM adaptation, likely drive its improved robustness to disfluency.


\Observation{10}{Training objectives shape capability profiles.}
Within the Qwen3-Omni family, the thinking and captioner variants are trained by Qwen3-Omni-Instruct for different objectives and exhibit different behaviors. The thinking model shows a \(6.8\%\) reduction in interquartile range, indicating more consistent performance with explicit thinking. The captioner achieves stronger speech recognition, with a \(2\) points improvement under content insertion, but performs worse on long context. These findings suggest that robustness to different disfluencies requires different capabilities, motivating disfluency-specific training objectives.

\section{Further Discussion}
\begin{figure*}[htbp]
    \centering
    \includegraphics[width=0.9\linewidth]{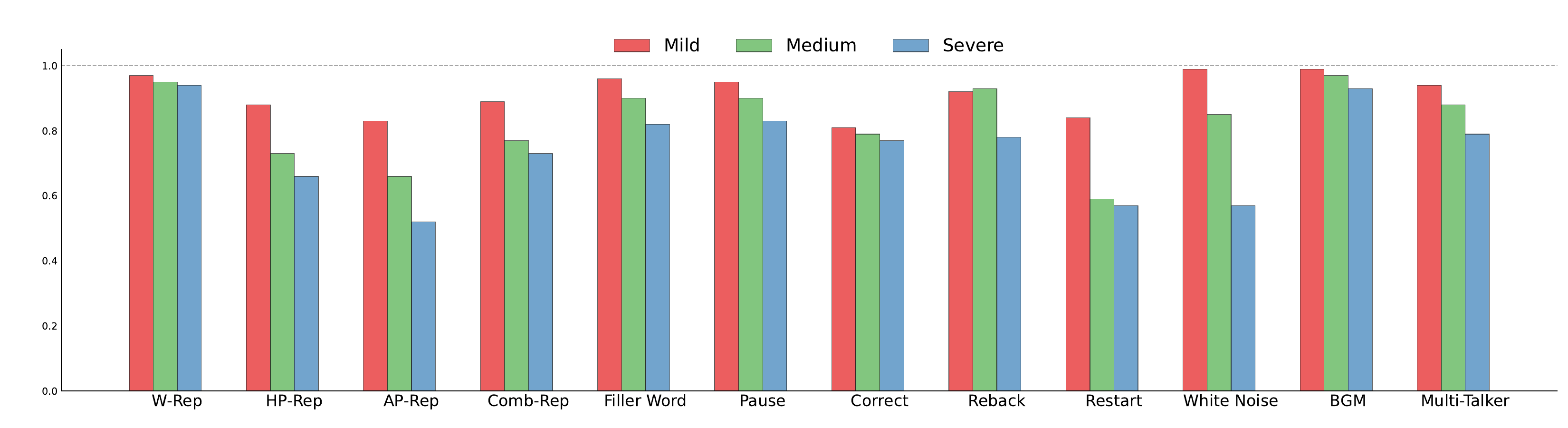}
    \caption{Average performance of Speech-LLMs across three disfluency severity levels. There are consistent stepwise declines for most types. (W-Rep: word repetition, HP-Rep: head phoneme repetition, AP-Rep: any phoneme repetition, Comb-Rep: combined repetition)}
    \label{fig:level}
    \vspace{-1.5em}
\end{figure*}

\begin{figure}[t]
    \centering
    \includegraphics[width=0.85\linewidth]{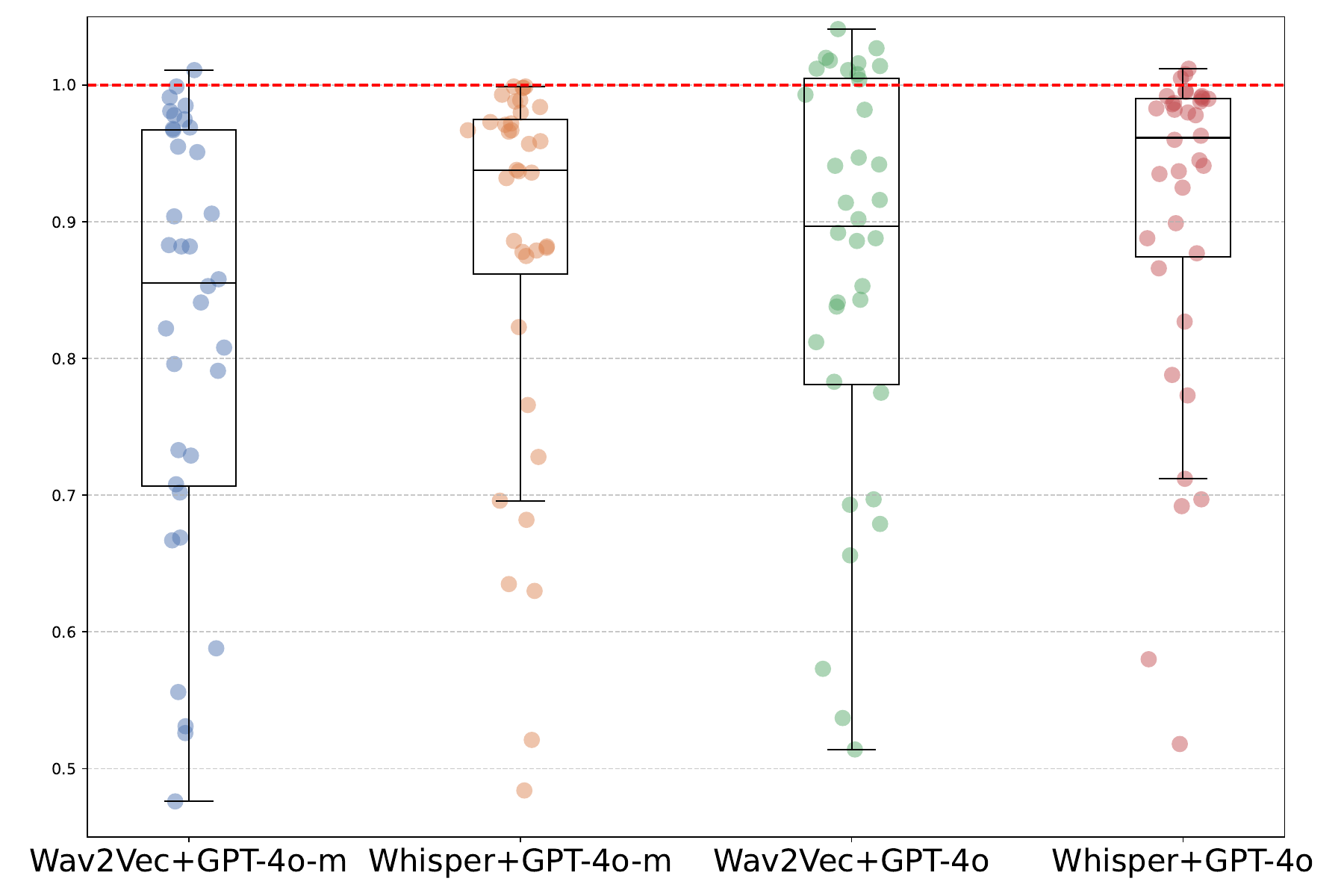}
    \caption{Performance of combinations of ASR models and LLMs. Each point denotes a score in a specific disfluency type. Whiskers show the range of non-outlier data. The box shows the interquartile range from \(25\%\) to \(75\%\). The line inside the box marks the median.}
    \label{fig:asr}
    \vspace{-1.5em}
\end{figure}

We further explore the underlying factors of robustness to speech disfluency, and the results are shown in Table~\ref{tab:stutter}, Figure~\ref{fig:level}, and Figure~\ref{fig:asr}.
\paragraph{Decomposing Repetition Disfluencies}
To locate the source of the degradation under repetition, we decompose it into four subtypes: word, head-phoneme, any-phoneme, and combined repetition.

\Observation{11}{Speech-LLMs always ignore the semantic modeling in phonemes.}
\begin{table}[t!]
\centering
\renewcommand{\arraystretch}{0.95}
\resizebox{0.48\textwidth}{!}{
\begin{tabular}{lccccc}
\toprule
\textbf{Model} & \textbf{W} & \textbf{HP} & \textbf{AP} & \textbf{C} & \textbf{Average}\\
\midrule
\rowcolor{gray!15} \multicolumn{6}{c}{\textbf{LLM Scale$\leq$ 3B}} \\
Mini-Omni & 0.84 & 0.55 & 0.44 & 0.60 & 0.61\\
Mini-Omni2 & 0.92 & 0.57 & 0.50 & 0.64 & 0.66\\
LLaMA2-0.5B & 0.93 & 0.61 & 0.58 & 0.63 & 0.69\\
VocalNet-1B & 0.93 & 0.60 & 0.53 & 0.64 & 0.68\\
LLaMA-Omni2-1.5B & 0.92 & 0.70 & 0.57 & 0.68 & 0.72\\
LLaMA-Omni2-3B & 0.97 & 0.77 & 0.67 & 0.79 & 0.80\\
Qwen-Omni-3B & 0.95 & 0.73 & 0.62 & 0.75 & 0.76\\
\midrule
\rowcolor{gray!15} \multicolumn{6}{c}{\textbf{3B $<$ LLM Scale $\leq$ 8B}} \\
Qwen2-Audio-7B & 0.81 & 0.49 & 0.43 & 0.53 & 0.57\\
Qwen-Omni-7B & 0.99 & 0.80 & 0.75 & 0.85 & 0.85\\
Kimi-Audio & 1.00 & 0.85 & 0.80 & 0.87 & 0.88\\
LLaMA-Omni2-7B & 0.94 & 0.73 & 0.65 & 0.76 & 0.77\\
VocalNet-qwen25-7B & 0.99 & 0.78 & 0.70 & 0.80 & 0.82\\
MiniCPM-o-2.6 & 0.98 & 0.75 & 0.66 & 0.79 & 0.80\\
Baichuan-Omni-1.5 & 0.95 & 0.70 & 0.65 & 0.75 & 0.76\\
Audio-Flamigo-3 &0.98 &0.75 &0.73 &0.80 &0.81\\
VocalNet-8B & 0.93 & 0.67 & 0.61 & 0.69 & 0.73\\
\midrule
\rowcolor{gray!15} \multicolumn{6}{c}{\textbf{LLM Scale $>$ 8B and Closed-Source}} \\
GLM-4-Voice& 0.92& 0.68& 0.57& 0.72& 0.72\\
LLaMA-Omni2-14B & 0.96 & 0.79 & 0.70 & 0.80 & 0.81\\
Qwen3-Omni-Instruct &1.00 &0.82 &0.71 &0.85 &0.85 \\
Qwen3-Omni-Captioner &1.02 &0.74& 0.78 &0.87&0.85\\
Qwen3-Omni-Thinking &0.99 &0.89 &0.85 &0.91&0.91\\
GPT-4o-audio & 1.01 & 0.89 & 0.84 & 0.94 & 0.92\\
\midrule
\textbf{Overall Average} & 0.95 & 0.72 & 0.65 & 0.76 & 0.77\\
\bottomrule
\end{tabular}
}
\caption{Performance comparison of various repetitions. 
(W: word repetition, HP: head phoneme repetition, AP: any phoneme repetition, C: combined repetition)}
\label{tab:stutter}
\vspace{-1.5em}
\end{table}
 
As shown in Table~\ref{tab:stutter}, word repetition has only a minor effect, and performance remains close to that in the normal setting (\(0.95\)). In contrast, phoneme-level repetition, especially any-phoneme repetition, produces a pronounced decrease in accuracy, yielding a \(28\%\) drop relative to the word-level case on average. These results suggest that current speech-LLMs can often preserve intended semantics when repetition occurs at the word level, but they lack mechanisms that normalize fine-grained phoneme repetitions into coherent semantic representations. These findings motivate the expansion of the coverage of disfluency and the improvement of accessibility for speakers with repetition or other speech disorders.

\Observation{12}{Medial-phoneme disfluencies are harder.}
As shown in Table~\ref{tab:stutter}, repetitions at medial phoneme positions perform worse than repetitions of the head phoneme, with an average degradation of about \(10\%\). Repeating the head phoneme preserves the fluent semantics of the remainder of the word, allowing models to ignore the leading repetition. By contrast, medial repetitions disrupt the internal structure and are harder to normalize. This suggests that models exhibit partial capability to onset-level phoneme disfluencies but remain insufficient for within-word repetitions.

\paragraph{Fine-Grained Severity Effects across Disfluencies}
To fully investigate the causes of degradation, we introduce three severity levels: mild, medium, and severe.

\Observation{13}{Many models can process the disfluency at an easy level, but cannot do it at a hard level.}
As shown in Figure~\ref{fig:level}, these levels produce clear stepwise performance gaps for most disfluency types, especially for repetition, pause, and white noise, which is not fully aligned with the trends in the main results. This pattern indicates that some models cope at the medium level but fail as severity increases, suggesting only partial rather than genuine stability. These findings imply that improving model robustness to disfluency requires severity-specific strategies, particularly for sharp stepwise degradation.

\Observation{14}{Reback exposes limits in long-context modeling}
Also in Figure~\ref{fig:level}, correction exerts relatively minor effects on semantics, whereas reback imposes a strong impact, especially under severe conditions, dropping from \(0.92\) to \(0.77\). This can be attributed to the fact that model capability declines substantially with longer context, and crucial information at the beginning of the input is more likely to be forgotten. These involve existing models that should pay more attention to long-context modeling. 

\paragraph{Verifying the Contribution of the Speech Recognition and Complex Reasoning.}
Building on these insights, we ask whether the capability of speech recognition and complex reasoning is really effective. We vary system design choices in the ASR and LLM components while holding all other factors constant. Specifically, we use Wav2Vec-base-960h and Whisper-large-v3 as ASR models, and GPT-4o mini and GPT-4o as LLMs, with the latter in each pair being stronger.

\Observation{15}{The combination of high-accuracy recognition and strong reasoning generally enhances overall performance.}
 Figure~\ref{fig:asr} yields two findings. First, stronger ASR models deliver more stable performance, with a \(52\%\) reduction in interquartile range. Cleaner transcriptions reduce recognition errors that would otherwise magnify the impact of disfluencies. Second, stronger LLMs achieve higher overall accuracy, with improvements exceeding \(4\%\). Better reasoning helps normalize disfluent inputs and recover the intended meaning. These results highlight the complementary roles of ASR accuracy and LLM reasoning capacity, suggesting that balance in both components is essential for strong handling of disfluencies in speech-LLM.
\section{Conclusions}
To advance the evaluation of Speech-LLMs under diverse disfluency conditions, we introduce \textsc{VocalBench-DF}, a novel and systematic benchmark for speech-disfluency assessment. The framework spans multi-dimensional disfluency categories with an emphasis on core speech characteristics. In a comprehensive study of 22 prominent Speech-LLMs, we find that current systems struggle to handle disfluencies effectively. Further analysis identifies phoneme disfluencies and long-context modeling as primary obstacles, with increasing disfluency levels causing substantial performance degradation. Nevertheless, integrating strong speech recognition with capable complex reasoning underscores the critical roles of strong components and process pipelines in robustness to disfluency. These findings chart key pathways toward targeted improvements and generalization under disfluent conditions.

\newpage
\section*{Limitations}
Our study has several limitations. First, it does not encompass the full spectrum of disfluency phenomena. Articulatory disorders and phoneme confusions, such as the contrast between alveolar and velar nasals, are not modeled. Second, the speech material is synthetically generated, which may fail to capture the full variability of spontaneous speech, background noise, and channel conditions. In the future, we will broaden disfluency coverage to include articulatory disorders and fine-grained phoneme confusions, and shift from synthetic to spontaneous speech under diverse noise/channel conditions.
\section*{Ethic Consideration}
We use OpenAudioBench as the source data and generate speech with CosyVoice, both publicly released resources. We follow the licenses and usage guidelines of these projects. The generated speech mirrors the textual prompts and contains no personal data. We screen content to avoid toxic, hateful, sexual, or otherwise unsafe material. All evaluated models are publicly available and accessed through their official interfaces. Human evaluation was conducted by college student raters. Participation did not require disclosure of personal characteristics, and no personally identifiable information was collected. Given these conditions, we do not identify additional ethical risks beyond those acknowledged by the creators of the datasets and models. Furthermore,  we leveraged large language models as writing assistants for tasks such as rephrasing sentences, improving grammatical flow, and refining technical descriptions for clarity.

\newpage
\bibliography{anthology,custom}
\bibliographystyle{acl_natbib}
\newpage
\appendix

\label{sec:appendix}

\section{The details on severity levels.}
\begin{table}[htbp]
  \centering
  \setlength{\tabcolsep}{5pt}
  \renewcommand{\arraystretch}{1.15}
  \resizebox{0.45\textwidth}{!}{%
  \begin{tabular}{@{}l l c c c@{}}
    \toprule
    \multirow{2}{*}{\textbf{Type}} & \multirow{2}{*}{\textbf{Parameter}} 
      & \multicolumn{3}{c}{\textbf{Severity}} \\
    \cmidrule(lr){3-5}
      &  & \textbf{Mild} & \textbf{Medium} & \textbf{Severe} \\
    \midrule
    \multirow{4}{*}{Rep} 
      & Probability         & 0.6 & 0.7 & 0.8 \\
      & Max frequency       & 2   & 3   & 4   \\
      & Repetition span     & 1–5 & 1–7 & 1–9 \\
      & Max relative span   & 1.5 & 2.0 & 2.5 \\
    \addlinespace[2pt]
    \multirow{3}{*}{Filler} 
      & Probability         & 0.1 & 0.2  & 0.35 \\
      & Max frequency       & 2   & 4    & 5    \\
      & Min frequency       & 1   & 2    & 3    \\
    \addlinespace[2pt]
    \multirow{3}{*}{Pause} 
      & Probability         & 0.3 & 0.5  & 0.7 \\
      & Max duration (s)    & 0.5 & 0.7  & 0.9 \\
      & Min duration (s)    & 0.1 & 0.2  & 0.3 \\
    \addlinespace[2pt]
    Correct  & Question count & 1    & 2    & 3    \\
    \addlinespace[2pt]
    Reback   & Duration (s)   & 52.2 & 67.6  & 94.5    \\
    \addlinespace[2pt]
    Restart  & Duration (s)   & 28.4    & 45.1    & 53.9    \\
    \addlinespace[2pt]
    W--N     & Level (dB)     & -35  & -25  & -15  \\
    \addlinespace[2pt]
    BGM      & Level (dB)     & -20  & -15  & -10  \\
    \addlinespace[2pt]
    M--T     & Level (dB)     & -20  & -15  & -10  \\
    \bottomrule
  \end{tabular}}
  \caption{Details of severity settings in our dataset. \\ \textbf{Type Abbrev. legend:} Rep = Repetition, W-N = White Noise, M-T = Multi-Talker.}
  \label{tab:severity_detail}
\end{table}

The severity of repetition, filler words, and speech pauses is primarily distinguished by the frequency and duration of the disfluent segments. For correction, reback, and restart, severity is determined by the length of task-irrelevant content introduced. In the case of white noise, background music, and multi-talker interference, severity correlates with the volume level of environmental sounds. The details are shown in Table~\ref{tab:severity_detail}.

\section{Further Experimental Results}

\begin{table}[htbp]
\centering
\setlength{\tabcolsep}{4pt}
\resizebox{0.35\textwidth}{!}{%
\begin{tabular}{l|ccccc}
\toprule
\textbf{Type} & \textbf{Seve.} & { WGm } & { WsGm } & { WG } & { WsG } \\
\midrule
\multirow{1}{*}{Nor} & Nor & 54.65 & 57.00 & 58.13 & 60.70 \\
\addlinespace[2pt]
\multirow{3}{*}{W\_Rep} & Mi. & 0.98 & 0.98 & 1.01 & 0.99 \\
 & Me. & 0.97 & 0.97 & 0.99 & 0.99 \\
 & Se. & 0.97 & 0.96 & 1.00 & 0.98 \\
\addlinespace[2pt]
\multirow{3}{*}{HP\_Rep} & Mi. & 0.88 & 0.88 & 0.94 & 0.90 \\
 & Me. & 0.80 & 0.73 & 0.85 & 0.79 \\
 & Se. & 0.71 & 0.63 & 0.81 & 0.71 \\
\addlinespace[2pt]
\multirow{3}{*}{AP\_Rep} & Mi. & 0.86 & 0.82 & 0.89 & 0.87 \\
 & Me. & 0.73 & 0.63 & 0.77 & 0.69 \\
 & Se. & 0.59 & 0.48 & 0.68 & 0.58 \\
\addlinespace[2pt]
\multirow{3}{*}{C\_Rep} & Mi. & 0.90 & 0.89 & 0.94 & 0.92 \\
 & Me. & 0.84 & 0.77 & 0.84 & 0.83 \\
 & Se. & 0.73 & 0.70 & 0.90 & 0.77 \\
\addlinespace[2pt]
\multirow{3}{*}{Filler} & Mi. & 0.95 & 0.97 & 0.98 & 0.99 \\
 & Me. & 0.91 & 0.93 & 0.95 & 0.96 \\
 & Se. & 0.85 & 0.88 & 0.92 & 0.93 \\
\addlinespace[2pt]
\multirow{3}{*}{Pause} & Mi. & 0.82 & 0.99 & 0.89 & 0.99 \\
 & Me. & 0.70 & 0.94 & 0.78 & 0.94 \\
 & Se. & 0.56 & 0.87 & 0.66 & 0.88 \\
\addlinespace[2pt]
\multirow{3}{*}{Correct} & Mi. & 0.98 & 0.97 & 1.03 & 0.99 \\
 & Me. & 0.97 & 0.97 & 1.02 & 0.99 \\
 & Se. & 0.95 & 0.97 & 1.01 & 0.99 \\
\addlinespace[2pt]
\multirow{3}{*}{Reback} & Mi. & 1.00 & 1.00 & 1.02 & 1.01 \\
 & Me. & 1.01 & 1.00 & 1.04 & 1.01 \\
 & Se. & 0.53 & 0.52 & 0.54 & 0.52 \\
\addlinespace[2pt]
\multirow{3}{*}{Restart} & Mi. & 0.98 & 0.98 & 1.01 & 1.00 \\
 & Me. & 0.99 & 0.99 & 1.02 & 0.99 \\
 & Se. & 0.98 & 0.97 & 1.01 & 0.98 \\
\addlinespace[2pt]
\multirow{3}{*}{W-N} & Mi. & 0.88 & 0.99 & 0.89 & 1.00 \\
 & Me. & 0.53 & 0.94 & 0.57 & 0.94 \\
 & Se. & 0.08 & 0.68 & 0.09 & 0.70 \\
\addlinespace[2pt]
\multirow{3}{*}{BGM} & Mi. & 0.88 & 1.00 & 0.91 & 0.99 \\
 & Me. & 0.79 & 0.99 & 0.84 & 0.98 \\
 & Se. & 0.67 & 0.96 & 0.69 & 0.97 \\
\addlinespace[2pt]
\multirow{3}{*}{M-T} & Mi. & 0.81 & 0.99 & 0.84 & 0.98 \\
 & Me. & 0.67 & 0.94 & 0.70 & 0.94 \\
 & Se. & 0.48 & 0.88 & 0.51 & 0.89 \\
\addlinespace[4pt]
\bottomrule
\end{tabular}}
\caption{The further results of ASR+LLM on various conditions. \\
\textbf{Type Abbrev. legend:} Nor = Normal, W\_Rep = Word Repetition, HP\_Rep = Head Phoneme Repetition, AP\_Rep = Any Phoneme Repetition, C\_Rep = Combined Repetition, W-N = White Noise, M-T = Multi-Talker. \\
\textbf{Severity Abbrev. legend:} Seve.=Severity, Mi.=Mild, Me.=Medium, Se.=Severe. \\
\textbf{Model Abbrev. legend:} \texttt{WGm}=Wav2Vec-GPT-4o-mini, \texttt{WsGm}=Whisper-GPT-4o-mini, \texttt{WG}=Wav2Vec-GPT-4o, \texttt{WsG}=Whisper-GPT-4o.}
\label{Tabl:fur_result_gpt}
\end{table}

To further analyze model performance, we report the complete results across disfluency types and difficulty levels in Tables~\ref{Tabl:fur_result_gpt} and~\ref{Tabl:fur_result}. There are several conclusions that can be obtained:

\begin{itemize}[itemsep=0pt, topsep=0pt, parsep=0pt]
    \item Model performance spans a wide range. Mini-Omni has the lowest performance, whereas Kimi-Audio attains the highest, even surpassing GPT-4o-Audio. These results suggest that carefully engineered architectures and training pipelines can substantially improve robustness to disfluency.
    \item Despite acceptable outcomes under mild disfluency, most systems degrade sharply as severity increases, indicating that high-severity conditions remain difficult to handle.
    \item Across disfluency types, phoneme-level repetition is consistently the most challenging at all three severity levels, reflecting a limited capacity to reconstruct meaning under phonemic disruptions.
    \item At the utterance level, models are generally competent in mild settings but deteriorate in severe ones, underscoring long-context modeling as a continuing bottleneck.
    \item Overall, achieving stable performance under high-severity disfluency remains an open research problem.
\end{itemize}


\begin{table*}[htbp]
\centering

\setlength{\tabcolsep}{4pt}
\resizebox{\textwidth}{!}{%
\begin{tabular}{l|ccccccccccccccccccccccc}
\toprule
\textbf{Type} & \textbf{Seve.} & { MO } & { M2 } & { L05 } & { V1 } & { L15 } & { L3 } & { Q3 } & { Q2 } & { Q7 } & { KA } & { L7 } & { VQ } & { MC } & { BC } & { AF3 } & { V8 } & { GV } & { L14 } & { Q3I } & { Q3C } & { Q3T } & { G4oA } \\
\midrule
\multirow{1}{*}{Nor} & Nor & 11.26 & 13.17 & 14.87 & 29.83 & 27.48 & 31.39 & 35.22 & 30.87 & 41.61 & 45.52 & 36.96 & 39.22 & 46.57 & 41.57 & 36.09 & 43.52 & 38.35 & 44.26 & 53.52 & 51.17 & 60.65 & 61.74 \\
\addlinespace[2pt]
\multirow{3}{*}{W\_Rep} & Mi. & 0.95 & 0.96 & 1.01 & 0.96 & 0.94 & 0.99 & 0.99 & 0.90 & 1.01 & 1.02 & 0.97 & 0.99 & 0.98 & 0.96 & 1.00 & 0.97 & 0.94 & 0.97 & - & - & - & 0.98 \\
 & Me. & 0.84 & 0.92 & 0.93 & 0.93 & 0.92 & 0.97 & 0.95 & 0.81 & 0.99 & 1.00 & 0.94 & 0.99 & 0.98 & 0.95 & 0.98 & 0.93 & 0.92 & 0.96 & 1.02 & 1.00 & 0.99 & 1.01 \\
 & Se. & 0.82 & 0.84 & 0.89 & 0.91 & 0.92 & 0.95 & 0.99 & 0.75 & 0.98 & 1.03 & 0.92 & 0.99 & 0.97 & 0.93 & 0.96 & 0.92 & 0.90 & 0.95 & - & - & - & 1.01 \\
\addlinespace[2pt]
\multirow{3}{*}{HP\_Rep} & Mi. & 0.75 & 0.73 & 0.78 & 0.80 & 0.81 & 0.86 & 0.91 & 0.76 & 0.91 & 0.95 & 0.89 & 0.91 & 0.90 & 0.85 & 0.91 & 0.85 & 0.85 & 0.89 & - & - & - & 0.98 \\
 & Me. & 0.55 & 0.57 & 0.61 & 0.60 & 0.70 & 0.77 & 0.73 & 0.49 & 0.80 & 0.85 & 0.73 & 0.78 & 0.75 & 0.70 & 0.75 & 0.67 & 0.68 & 0.79 & 0.74 & 0.82 & 0.89 & 0.99 \\
 & Se. & 0.53 & 0.52 & 0.57 & 0.49 & 0.64 & 0.69 & 0.59 & 0.40 & 0.72 & 0.82 & 0.70 & 0.71 & 0.68 & 0.62 & 0.69 & 0.55 & 0.59 & 0.72 & - & - & - & 0.94 \\
\addlinespace[2pt]
\multirow{3}{*}{AP\_Rep} & Mi. & 0.70 & 0.73 & 0.78 & 0.78 & 0.77 & 0.81 & 0.82 & 0.68 & 0.89 & 0.88 & 0.83 & 0.86 & 0.83 & 0.79 & 0.85 & 0.81 & 0.79 & 0.86 & - & - & - & 0.95 \\
 & Me. & 0.44 & 0.50 & 0.58 & 0.53 & 0.57 & 0.67 & 0.62 & 0.43 & 0.75 & 0.80 & 0.65 & 0.70 & 0.66 & 0.65 & 0.73 & 0.61 & 0.57 & 0.70 & 0.78 & 0.71 & 0.85 & 0.84 \\
 & Se. & 0.41 & 0.43 & 0.41 & 0.36 & 0.44 & 0.54 & 0.44 & 0.28 & 0.60 & 0.66 & 0.52 & 0.52 & 0.50 & 0.48 & 0.54 & 0.43 & 0.44 & 0.56 & - & - & - & 0.98 \\
\addlinespace[2pt]
\multirow{3}{*}{C\_Rep} & Mi. & 0.77 & 0.82 & 0.81 & 0.86 & 0.85 & 0.91 & 0.91 & 0.76 & 0.93 & 0.93 & 0.89 & 0.91 & 0.92 & 0.87 & 0.92 & 0.85 & 0.86 & 0.91 & - & - & - & 1.00 \\
 & Me. & 0.60 & 0.64 & 0.63 & 0.64 & 0.68 & 0.79 & 0.75 & 0.53 & 0.85 & 0.87 & 0.76 & 0.80 & 0.79 & 0.75 & 0.80 & 0.69 & 0.72 & 0.80 & 0.87 & 0.85 & 0.91 & 0.99 \\
 & Se. & 0.51 & 0.58 & 0.63 & 0.60 & 0.62 & 0.75 & 1.03 & 0.41 & 0.78 & 0.87 & 0.71 & 0.77 & 0.73 & 0.66 & 0.73 & 0.62 & 0.61 & 0.76 & - & - & - & 0.94 \\
\addlinespace[2pt]
\multirow{3}{*}{Filler} & Mi. & 0.82 & 0.94 & 0.97 & 0.92 & 0.96 & 0.99 & 0.98 & 0.87 & 0.98 & 0.99 & 0.96 & 0.98 & 0.94 & 0.95 & 0.94 & 0.92 & 0.91 & 0.97 & - & - & - & 1.00 \\
 & Me. & 0.70 & 0.78 & 0.83 & 0.86 & 0.91 & 0.94 & 0.92 & 0.74 & 0.94 & 0.97 & 0.91 & 0.93 & 0.92 & 0.89 & 0.87 & 0.86 & 0.81 & 0.93 & 0.95 & 0.95 & 0.97 & 0.99 \\
 & Se. & 0.62 & 0.73 & 0.82 & 0.74 & 0.86 & 0.89 & 0.79 & 0.55 & 0.84 & 0.95 & 0.90 & 0.85 & 0.86 & 0.81 & 0.78 & 0.77 & 0.74 & 0.90 & - & - & - & 0.99 \\
\addlinespace[2pt]
\multirow{3}{*}{Pause} & Mi. & 0.92 & 0.95 & 0.92 & 0.92 & 0.91 & 0.95 & 0.98 & 0.92 & 0.99 & 1.00 & 0.92 & 0.92 & 0.97 & 0.96 & 0.93 & 0.91 & 0.92 & 0.92 & - & - & - & 1.00 \\
 & Me. & 0.81 & 0.83 & 0.89 & 0.88 & 0.86 & 0.89 & 0.93 & 0.81 & 0.95 & 0.99 & 0.86 & 0.90 & 0.94 & 0.89 & 0.87 & 0.86 & 0.84 & 0.85 & 0.98 & 0.96 & 0.99 & 0.99 \\
 & Se. & 0.74 & 0.80 & 0.78 & 0.79 & 0.75 & 0.81 & 0.86 & 0.74 & 0.91 & 0.98 & 0.76 & 0.83 & 0.88 & 0.79 & 0.78 & 0.79 & 0.72 & 0.79 & - & - & - & 0.94 \\
\addlinespace[2pt]
\multirow{3}{*}{Correct} & Mi. & 0.28 & 0.31 & 0.47 & 0.65 & 0.63 & 0.73 & 0.90 & 0.39 & 0.93 & 1.00 & 0.81 & 0.98 & 0.83 & 0.81 & 0.80 & 0.89 & 0.79 & 0.87 & - & - & - & 0.84 \\
 & Me. & 0.19 & 0.18 & 0.41 & 0.58 & 0.68 & 0.74 & 0.86 & 0.43 & 0.91 & 0.99 & 0.83 & 0.98 & 0.85 & 0.75 & 0.68 & 0.88 & 0.78 & 0.88 & 0.95 & 0.98 & 0.89 & 0.98 \\
 & Se. & 0.22 & 0.16 & 0.45 & 0.53 & 0.72 & 0.79 & 0.87 & 0.48 & 0.92 & 0.94 & 0.86 & 0.97 & 0.77 & 0.68 & 0.65 & 0.89 & 0.78 & 0.87& - & - & -  & 1.00 \\
\addlinespace[2pt]
\multirow{3}{*}{Reback} & Mi. & 0.59 & 0.77 & 0.94 & 0.76 & 0.96 & 1.02 & 0.94 & 0.79 & 0.96 & 1.03 & 0.97 & 0.99 & 0.98 & 0.97 & 0.87 & 0.78 & 0.85 & 0.98 & - & - & - & 0.94 \\
 & Me. & 0.60 & 0.78 & 0.93 & 0.75 & 0.91 & 0.99 & 0.98 & 0.76 & 0.99 & 1.05 & 0.95 & 0.97 & 0.99 & 0.97 & 0.97 & 0.69 & 0.93 & 0.96 & 0.80 & 0.99 & 0.94 & 0.20 \\
 & Se. & 0.58 & 0.79 & 0.88 & 0.71 & 0.94 & 0.95 & 0.91 & 0.46 & 0.50 & 0.51 & 0.95 & 0.94 & 0.94 & 0.50 & 0.71 & 0.68 & 0.71 & 0.94 & - & - & - & 0.02 \\
\addlinespace[2pt]
\multirow{3}{*}{Restart} & Mi. & 0.49 & 0.58 & 0.56 & 0.80 & 0.66 & 0.69 & 0.91 & 0.66 & 0.94 & 1.00 & 0.70 & 0.99 & 0.90 & 0.90 & 0.87 & 0.82 & 0.82 & 0.77 & - & - & - & 1.00 \\
 & Me. & 0.23 & 0.25 & 0.25 & 0.22 & 0.23 & 0.20 & 0.90 & 0.17 & 0.95 & 1.03 & 0.19 & 0.22 & 0.92 & 0.90 & 0.89 & 0.21 & 0.87 & 0.19 & 0.19 & 0.29 & 0.28 & 0.96 \\
 & Se. & 0.21 & 0.25 & 0.23 & 0.16 & 0.17 & 0.17 & 0.90 & 0.13 & 0.94 & 1.01 & 0.16 & 0.17 & 0.92 & 0.89 & 0.88 & 0.16 & 0.89 & 0.16 & - & - & - & 0.94 \\
\addlinespace[2pt]
\multirow{3}{*}{W-N} & Mi. & 0.93 & 1.00 & 1.04 & 1.00 & 0.96 & 0.98 & 1.02 & 1.02 & 1.00 & 0.99 & 0.98 & 1.01 & 0.99 & 1.00 & 0.97 & 1.00 & 0.94 & 0.99 & - & - & - & 1.00 \\
 & Me. & 0.84 & 0.88 & 0.93 & 0.93 & 0.88 & 0.90 & 0.96 & 0.95 & 0.98 & 0.96 & 0.87 & 0.92 & 0.90 & 0.91 & 0.90 & 0.94 & 0.81 & 0.91 & 0.97 & 0.95 & 0.97 & 0.99 \\
 & Se. & 0.57 & 0.58 & 0.65 & 0.66 & 0.59 & 0.59 & 0.74 & 0.70 & 0.76 & 0.71 & 0.61 & 0.64 & 0.63 & 0.61 & 0.54 & 0.66 & 0.40 & 0.61 & - & - & - & 0.99 \\
\addlinespace[2pt]
\multirow{3}{*}{BGM} & Mi. & 0.95 & 1.02 & 1.01 & 0.98 & 0.97 & 0.99 & 1.00 & 1.00 & 1.00 & 0.99 & 0.98 & 0.97 & 0.97 & 0.98 & 1.00 & 1.00 & 0.94 & 0.99 & - & - & - & 0.94 \\
 & Me. & 0.96 & 0.98 & 0.99 & 0.97 & 0.95 & 0.95 & 0.98 & 1.00 & 0.99 & 0.99 & 0.98 & 0.94 & 0.96 & 0.98 & 0.97 & 0.97 & 0.92 & 0.97 & 1.00 & 0.98 & 0.99 & 0.20 \\
 & Se. & 0.82 & 0.91 & 0.98 & 0.95 & 0.90 & 0.94 & 0.96 & 0.96 & 0.96 & 0.96 & 0.93 & 0.92 & 0.91 & 0.96 & 0.90 & 0.95 & 0.86 & 0.93 & - & - & - & 0.02 \\
\addlinespace[2pt]
\multirow{3}{*}{M-T} & Mi. & 0.90 & 0.91 & 0.88 & 0.92 & 0.84 & 0.91 & 0.97 & 0.97 & 0.99 & 0.99 & 0.89 & 0.92 & 0.97 & 0.94 & 0.95 & 0.90 & 0.94 & 0.91 & - & - & - & 1.00 \\
 & Me. & 0.82 & 0.90 & 0.77 & 0.84 & 0.75 & 0.81 & 0.95 & 0.88 & 0.94 & 0.98 & 0.82 & 0.86 & 0.92 & 0.92 & 0.89 & 0.84 & 0.89 & 0.81 & 0.94 & 0.94 & 0.96  & 0.97 \\
 & Se. & 0.80 & 0.81 & 0.69 & 0.73 & 0.65 & 0.72 & 0.84 & 0.82 & 0.86 & 0.94 & 0.67 & 0.76 & 0.86 & 0.85 & 0.79 & 0.73 & 0.82 & 0.69 & - & - & - & 0.85 \\
\addlinespace[4pt]
\bottomrule
\end{tabular}}
\caption{
The further results on various conditions of speech-LLMs.\\
\textbf{Type Abbrev. legend:} Nor = Normal, W\_Rep = Word Repetition, HP\_Rep = Head Phoneme Repetition, AP\_Rep = Any Phoneme Repetition, C\_Rep = Combined Repetition, W-N = White Noise, M-T = Multi-Talker. \\
\textbf{Severity Abbrev. legend:} Seve.=Severity, Mi.=Mild, Me.=Medium, Se.=Severe. \\
\textbf{Model Abbrev. legend:} \texttt{MO}=Mini-Omni; \texttt{M2}=Mini-Omni2; \texttt{L05}=LLaMA-Omni2-0.5b; \texttt{V1}=VocalNet-1B; \texttt{L15}=LLaMA-Omni2-1.5b; \texttt{L3}=LLaMA-Omni2-3b; \texttt{Q3}=Qwen2.5-Omni-3B; \texttt{Q2}=Qwen2-Audio-7B; \texttt{Q7}=Qwen2.5-Omni-7B; \texttt{KA}=Kimi-audio; \texttt{L7}=LLaMA-Omni2-7b; \texttt{VQ}=VocalNet-qwen25-7B; \texttt{MC}=MiniCPM-o-2.6; \texttt{BC}=Baichuan-omni-1.5; \texttt{AF3}=Audio-Flamingo-3; \texttt{V8}=VocalNet-8B; \texttt{GV}=GLM-4-Voice; \texttt{L14}=LLaMA-Omni2-14b; \texttt{Q3I}=Qwen3-Omni-Instruct; \texttt{Q3C}=Qwen3-Omni-Captioner; \texttt{Q3T}=Qwen3-Omni-Thinking; \texttt{G4oA}=GPT-4o-audio.}
\label{Tabl:fur_result}
\end{table*}

\section{The details of Models}
\label{app-sec:models}
For further analysis of the deep explanation of the various performances under disfluency conditions, we further describe the models in table~\ref{tab:model_detail}. As shown in the description, models that invest heavily in the speech encoder consistently achieve stronger results, indicating the encoder’s central effectiveness. The LLM also plays an important role in query understanding and intent reasoning. Beyond simply using a powerful encoder or a superior LLM, the top-performing systems are characterized by carefully designed, multi-stage alignment pipelines, which appear to drive much of the performance gains in the training pipeline for alignment.
\newcolumntype{Y}{>{\footnotesize\centering\arraybackslash}X}  
\newcolumntype{L}{>{\small\bfseries}l}                         

\begin{table*}[t]
\centering
\small
\renewcommand{\arraystretch}{1.35}
\setlength{\tabcolsep}{1pt}
\begin{tabularx}{\textwidth}{@{} >{\bfseries}l *{3}{Y} @{}}
\toprule
\textbf{Model} & \textbf{Speech Encoder} & \textbf{LLM Backbone} & \textbf{Train Pipeline} \\
\midrule
Mini-Omni
& Whisper-small + adapter
& Qwen2-0.5B
& Modality alignment + Adaptation training + Multimodal fine-tuning \\

Mini-Omni2
& Whisper-small + adapter
& Qwen2-0.5B
& Encoder adaptation + alignment + post-training \\

LLaMA-Omni2-series
& Whisper-large-v3 + adapter
& Qwen2.5 series
& Separate training: speech adapter + LLM; TTS LM; Gate fusion; S2S TTS \\

VocalNet-series
& Whisper-large-v3 + adapter
& LLaMA-3.2-1B-Instruct; Qwen2.5-7B; LLaMA-3.1-8B-Instruct
& Two phase: Downsample adaptor + LLM; Speech projector + Speech decoder \\ 

Qwen2.5-Omni-series
& Initialized from Whisper-large-v3
& Qwen2.5 series
& Three-stage pre-training + Talker post-training \\

Qwen2-Audio-7B
& Initialized from Whisper-large-v3
& Qwen-7B
& Pre-training + SFT + DPO \\

Kimi-Audio
& Audio-Tokenizer: a hybrid audio tokenization strategy integrating discrete semantic tokens and complementary continuous acoustic vectors
& Initialized from Qwen2.5-7B
& Pre-training + SFT + Detokenizer training \\

MiniCPM-o-2.6
& Whisper-medium-300M
& Qwen2.5-7B-Instruct
& Pre-training + Omni SFT + RLAIF \\

Baichuan-Omni-1.5
& Baichuan-Audio-Tokenizer based on residual vector quantization and multi-objective training
& Qwen2.5-7B
& Image--Text pre-train + Image--Audio--Text pre-train + Omni-modal pre-train + Multimodal SFT (freeze $\rightarrow$ activate audio head) \\

Audio-Flamingo-3
& AF-Whisper (Whisper-large-v3-based) + Adaptor
& Qwen2.5-7B
& Alignment pre-train + Encoder tuning + Full fine-tune + Reasoning extension + Chat SFT \\

GLM4-Voice
& GLM-4-Voice-Tokenizer: Whisper-large-v3 with an additional pooling layer and a vector quantization layer in the middle of the encoder
& GLM-4-9B
& Pre-training + SFT \\

Qwen3-omni-Instruct
& Audio Transformer(AuT) encoder scratch-trained on 20M\,h audio
& Qwen3-30B-A3B
& Three-stage pre-training + multi-stage post-training \\
\bottomrule
\end{tabularx}
\caption{Model details with speech-oriented elements.}
\label{tab:model_detail}
\end{table*}

\section{Case Study}
To clarify model performance across different disfluency conditions, we present examples in Tables~\ref{tab:case_start}--\ref{tab:case_final}. For each disfluency type, we include one successful and one failing case for the same set of models to allow direct comparison, highlighting the error patterns that differentiate success from failure.
\begin{table*}[t]
\centering
\small
\setlength{\tabcolsep}{8pt}
\renewcommand{\arraystretch}{1.15}

\begin{tabularx}{\textwidth}{>{\bfseries}p{0.20\textwidth} X}
\rowcolor[HTML]{343434}
\multicolumn{2}{c}{\color{white}\textbf{Word Repetition}}\\
\toprule
\rowcolor[HTML]{F0F0F0}\multicolumn{2}{c}{\textbf{Kimi-Audio} with True Answer}\\
Original Question & Which river is the longest in South America? \\
\midrule 
Ground Truth & Amazon \\
\midrule
Disturbed Question & Which-which river-river is the longest-longest in South America-america?\\
\midrule
Original Response & The Amazon River is the longest river in South America. It stretches for approximately 6,400 kilometers (4,000 miles) from its source in the Andes Mountains in Peru to its mouth in the Atlantic Ocean in Brazil. \\ \midrule
After & The Amazon River is the longest river in South America. It stretches for approximately 6,400 kilometers (4,000 miles) from its source in the Andes Mountains in Peru to its mouth in the Atlantic Ocean in Brazil. The Amazon River is not only the longest river in South America, but it is also the largest river in the world by discharge volume, accounting for about 15\% of the fresh water that flows into the oceans. The Amazon River basin covers an area of over 7 million square kilometers (2.7 million square miles), which is roughly the size of the United States. The river is home to a diverse array of plant and animal species, including over 2,000 species of fish, many of which are found nowhere else in the world. The Amazon River is also an important source of freshwater for millions of people in South America, providing drinking water, irrigation for agriculture, and hydroelectric power. \\ 
\midrule
\rowcolor[HTML]{F0F0F0}\multicolumn{2}{c}{\textbf{Kimi-Audio}  with Wrong Answer}\\
Original Question & What is the largest mamel? \\
\midrule 
Ground Truth & Blue Whale \\
\midrule
Disturbed Question & What-what is-is the largest-largest mamel-mamel?\\
\midrule
Original Response & The largest mammal is the blue whale (Balaenoptera musculus). On average, an adult blue whale can grow up to 82 feet (25 meters) in length and weigh around 150-170 tons (136,000-152,000 kilograms). However, the largest blue whale ever recorded was a female that was found in 1947 off the coast of Iceland, which measured around 108 feet (33 meters) in length and weighed an estimated 210 tons (182,000 kilograms). Blue whales are not only the largest mammals, but they are also the largest animals that have ever lived on Earth, even larger than the largest dinosaurs. They are found in all of the world's oceans and are known for their distinctive blue-gray color and their ability to produce extremely loud sounds, which can be heard for hundreds of miles. \\ \midrule
After  & The largest land mammal is the African elephant. On average, an adult male African elephant can weigh between 4,000-7,500 kg (8,800-16,500 lbs) and reach a height of 3-4 meters (10-13 feet) at the shoulder. However, some males have been known to reach weights of up to 10,000 kg (22,000 lbs) and heights of 4.5 meters (15 feet) or more. The African elephant is a herbivore and feeds on a variety of plants, including grasses, leaves, fruits, and bark. It is a social animal that lives in large family groups, called herds, which are led by a matriarch. Elephants are known for their intelligence, memory, and complex social behavior. It's worth noting that the largest land mammal of all time is the now-extinct Indricotherium, which lived during the Late Miocene epoch. It is estimated to have weighed around 15-20 tons (30,000-40,000 lbs) and reached lengths of up to 6 meters (20 feet). However, the African elephant is the largest land mammal that still exists today. \\
\bottomrule
\end{tabularx}
\caption{The case study of word repetition.}
\label{tab:case_start}
\end{table*}

\begin{table*}[t]
\centering
\small
\setlength{\tabcolsep}{8pt}
\renewcommand{\arraystretch}{1.15}

\begin{tabularx}{\textwidth}{>{\bfseries}p{0.20\textwidth} X}
\rowcolor[HTML]{343434}
\multicolumn{2}{c}{\color{white}\textbf{Head Phoneme Repetition}}\\
\toprule
\rowcolor[HTML]{F0F0F0}\multicolumn{2}{c}{\textbf{Qwen2.5-Omni-3B} with True Answer}\\
Original Question & What is the highest mountain peak in the contiguous United States? \\
\midrule 
Ground Truth & Whitney \\
\midrule
Disturbed Question & What is t-the h-h-h-highest mountain peak in the co-contiguous United S-states?\\
\midrule
Original Response & The highest mountain peak in the contiguous United States is Mount Whitney. It's located in California. It has a height of about 14, 505 feet. If you want to know more about Mount Whitney, like how to get there or some interesting facts about it, feel free to ask. \\ \midrule
After & Tahoe is not a mountain peak. It's a lake in the United States. The highest peak in the contiguous United States is Mount Whitney. It's in California.If you have any other questions about mountains or other things, feel free to ask. \\ \midrule
\rowcolor[HTML]{F0F0F0}\multicolumn{2}{c}{\textbf{Qwen2.5-Omni-3B}  with Wrong Answer}\\
Original Question & Which African country has the largest population? \\
\midrule 
Ground Truth & Nigeria \\
\midrule
Disturbed Question & Which A-african cou-country has the lar-largest p-population?\\
\midrule
Original Response & The country with the largest population in Africa is Nigeria. It has a very large and diverse population. If you want to know more about Nigeria or other African countries, feel free to ask. \\ \midrule
After  & Well, South Africa has the largest population among the African co - countries. It has a large number of people living there. If you want to know more about South Africa or other African co - countries, feel free to ask me. \\
\bottomrule
\end{tabularx}

\caption{The case study of head phoneme repetition.}
\label{tab:case_example}
\end{table*}

\begin{table*}[t]
\centering
\small
\setlength{\tabcolsep}{8pt}
\renewcommand{\arraystretch}{1.15}

\begin{tabularx}{\textwidth}{>{\bfseries}p{0.20\textwidth} X}
\rowcolor[HTML]{343434}
\multicolumn{2}{c}{\color{white}\textbf{Any Phoneme Repetition}}\\
\toprule
\rowcolor[HTML]{F0F0F0}\multicolumn{2}{c}{\textbf{Qwen2.5-Omni-7B} with True Answer}\\
Original Question & What is the highest mountain peak in North America? \\
\midrule 
Ground Truth & Denali \\
\midrule
Disturbed Question & What i-is the-he highest mount-ountain peak in North America-a?\\
\midrule
Original Response & The highest mountain peak in North America is Denali, formerly known as Mount McKinley. It's located in Alaska and has an elevation of about 20, 310 feet, 6, 190 meters. So, that's the answer. If you want to know more about Denali or other mountain peaks, feel free to ask. \\ \midrule
After & The highest mountain peak in North America is Denali, which is also known as Mount McKinley. Its elevation is about 20, 310 feet, 6, 190 meters. It's located in Alaska, United States. So, if you're interested in climbing or just learning about the highest mountains, Denali is a really cool place to know about. Do you have any other questions about mountains or North America? \\
\rowcolor[HTML]{F0F0F0}\multicolumn{2}{c}{\textbf{Qwen2.5-Omni-7B}  with Wrong Answer}\\
Original Question & What is the smallest country in the world? \\
\midrule 
Ground Truth & Vatican City \\
\midrule
Disturbed Question & Wha-what is-s the smallest-t count-country in the world?\\
\midrule
Original Response & Vatican City is the smallest country in the world. It's only about 0.44 square kilometers in size. It's located in Rome, Italy. So, if you're interested in small countries, Vatican City is a really interesting one. Do you want to know more about it?\\ \midrule
After  & Well, Monaco is often considered one of the smallest countries in the world. It's really tiny, with an area of about 2.02 square kilometers. But there are also some other small countries like Liechtenstein and San Marino. So, it's a bit of a debate. But Monaco is definitely one of the smallest. If you want to know more about these small countries, like their economies or cultures, just let me know.\\
\bottomrule
\end{tabularx}

\caption{The case study of any phoneme repetition.}
\label{tab:case_example}
\end{table*}

\begin{table*}[t]
\centering
\small
\setlength{\tabcolsep}{8pt}
\renewcommand{\arraystretch}{1.15}

\begin{tabularx}{\textwidth}{>{\bfseries}p{0.20\textwidth} X}
\rowcolor[HTML]{343434}
\multicolumn{2}{c}{\color{white}\textbf{Combined Repetition}}\\
\toprule
\rowcolor[HTML]{F0F0F0}\multicolumn{2}{c}{\textbf{Qwen2-Audio-7B} with True Answer}\\
Original Question & Which country borders Canada and Mexico? \\
\midrule 
Ground Truth & United States \\
\midrule
Disturbed Question & W-which co-country borders-rs Canada and Mexico-mexico?\\
\midrule
Original Response & The country that borders both Canada and Mexico is the United States of America. The border between the United States and Canada is the longest land border in the world, stretching for approximately 9,500 miles (15,000 kilometers). The border between the United States and Mexico is approximately 3,200 miles (5,100 kilometers) long.\\ \midrule
After & The state of Coahuila borders the United States to the north, with the Rio Grande (also known as the Rio Bravo) serving as the border. To the east, it shares a land border with the states of Nuevo León and Tamaulipas. To the south, it borders the states of Nuevo México and Sonora, and to the west, it is bordered by the state of Chihuahua.\\
\rowcolor[HTML]{F0F0F0}\multicolumn{2}{c}{\textbf{Qwen2-Audio-7B}  with Wrong Answer}\\
Original Question & Who was the first president of the United States? \\
\midrule 
Ground Truth & George Washington \\
\midrule
Disturbed Question & W-who was the first pr-pr-pr-president of t-the Un-united States?\\
\midrule
Original Response & The first president of the United States was George Washington.\\ \midrule
After & There is no such thing as the "united states". The United States of America is a federal republic composed of 50 states, each with its own government and laws. Please clarify your question or provide more information.\\
\bottomrule
\end{tabularx}

\caption{The case study of combined repetition.}
\label{tab:case_example}
\end{table*}
\begin{table*}[t]
\centering
\small
\setlength{\tabcolsep}{8pt}
\renewcommand{\arraystretch}{1.15}

\begin{tabularx}{\textwidth}{>{\bfseries}p{0.20\textwidth} X}
\rowcolor[HTML]{343434}
\multicolumn{2}{c}{\color{white}\textbf{Filler Words}}\\
\toprule
\rowcolor[HTML]{F0F0F0}\multicolumn{2}{c}{\textbf{Baichuan-Omni-1.5} with True Answer}\\
Original Question & What is the capital of Australia? \\
\midrule 
Ground Truth & Canberra \\
\midrule
Disturbed Question &What, like, is the capital, right, of Australia?\\
\midrule
Original Response & The capital city of Australia is Canberra.\\\midrule
After & The capital city of Australia is Canberra.\\\midrule
\rowcolor[HTML]{F0F0F0}\multicolumn{2}{c}{\textbf{Baichuan-Omni-1.5}  with Wrong Answer}\\
Original Question & What is the name of the largest desert in the world? \\
\midrule 
Ground Truth &  Antarctica\\
\midrule
Disturbed Question &What is the name of the largest, sort of, desert in the world, you know,?\\
\midrule
Original Response & The largest desert in the world is called Antarctica, although it's technically a continent and an ice desert rather than a hot desert like most people imagine when they think of deserts. If referring to traditional hot deserts, then the largest by area is the Sahara Desert located in North Africa.\\ \midrule
After & The largest hot desert in the world is called the Sahara Desert.\\
\bottomrule
\end{tabularx}

\caption{The case study of filler words.}
\label{tab:case_example}
\end{table*}

\begin{table*}[t]
\centering
\small
\setlength{\tabcolsep}{8pt}
\renewcommand{\arraystretch}{1.05}

\begin{tabularx}{\textwidth}{>{\bfseries}p{0.20\textwidth} X}
\rowcolor[HTML]{343434}
\multicolumn{2}{c}{\color{white}\textbf{Speech Pause}}\\
\toprule
\rowcolor[HTML]{F0F0F0}\multicolumn{2}{c}{\textbf{MiniCPM-o-2.6} with True Answer}\\
Original Question & What is the largest planet in our solar system? \\
\midrule 
Ground Truth &  Jupiter\\
\midrule
Disturbed Question & What \textit{---} is \textit{---}  the largest planet \textit{---} in our solar system?\\
\midrule
Original Response & The largest planet in our solar system is Jupiter. It's a gas giant with a diameter of about 143,000 kilometers (89,000 miles). That makes it more than twice the size of Earth\\\midrule
After & The largest planet in our solar system is Jupiter. It's a gas giant with a diameter of about 143,000 kilometers (89,000 miles), making it more than twice the size of Earth.\\\midrule
\rowcolor[HTML]{F0F0F0}\multicolumn{2}{c}{\textbf{MiniCPM-o-2.6} with Wrong Answer}\\
Original Question &  What is the bright object in the night sky that is not a star?\\
\midrule 
Ground Truth &  Moon\\
\midrule
Disturbed Question &What \textit{---} is the \textit{---} bright \textit{---} object \textit{---} in the night sky that is not a star?\\
\midrule
Original Response & The bright object in the night sky that is not a star could be several things. It might be: 1. The Moon, especially during its full phase when it's very luminous and visible at night. 2. A planet like Venus or Jupiter if they're particularly close to Earth and well-placed for viewing from our perspective on Earth. 3. An artificial satellite reflecting sunlight back towards us - these can sometimes appear quite bright as they orbit overhead. Without more specific details about where you are located and what time of year this observation was made (which would help determine which planets might be visible), I can't pinpoint exactly.\\ \midrule
After & The bright object in the night sky that is not a star appears to be Venus. It's one of the brightest objects visible in our night sky, often mistaken for an airplane or even a satellite due to its brilliance and steady appearance without twinkling like stars do.\\
\bottomrule
\end{tabularx}

\caption{The case study of combined repetition. \textit{---} denotes the pause in 0-1 seconds.}
\label{tab:case_example}
\end{table*}
\begin{table*}[t]
\centering
\small
\setlength{\tabcolsep}{8pt}
\renewcommand{\arraystretch}{1.15}

\begin{tabularx}{\textwidth}{>{\bfseries}p{0.20\textwidth} X}
\rowcolor[HTML]{343434}
\multicolumn{2}{c}{\color{white}\textbf{Correction}}\\
\toprule
\rowcolor[HTML]{F0F0F0}\multicolumn{2}{c}{\textbf{VocalNet-1B} with True Answer}\\
Original Question &  What is the largest planet in our solar system?\\
\midrule 
Ground Truth &  Jupiter\\
\midrule
Disturbed Question &What is the name of the test cricket venue in Manchester, England? where is tupac from originally? Oh no wait, I mean, What is the largest planet in our solar system?\\
\midrule
Original Response & The largest planet in our solar system is Jupiter.\\
\midrule 
After & The largest planet in our solar system is Jupiter.\\
\midrule 
\rowcolor[HTML]{F0F0F0}\multicolumn{2}{c}{\textbf{VocalNet-1B}  with Wrong Answer}\\
Original Question & Who was the first president of the United States? \\
\midrule 
Ground Truth & George Washington \\
\midrule
Disturbed Question & Which 19th century American poet wrote over 1,000 poems which were published between 1890 and 1945, after her death? What is the capital of France? Oh no wait, I mean, Who was the first president of the United States?\\
\midrule
Original Response & The first president of the United States was George Washington.\\
After & The capital of France is Paris.\\
\bottomrule
\end{tabularx}

\caption{The case study of correction.}
\label{tab:case_example}
\end{table*}

\begin{table*}[t]
\centering
\small
\setlength{\tabcolsep}{8pt}
\renewcommand{\arraystretch}{1.05}
\begin{tabularx}{\textwidth}{>{\bfseries}p{0.20\textwidth} X}
\rowcolor[HTML]{343434}
\multicolumn{2}{c}{\color{white}\textbf{Reback}}\\
\toprule
\rowcolor[HTML]{F0F0F0}\multicolumn{2}{c}{\textbf{VocalNet-8B} with True Answer}\\
Original Question & What is the currency used in Japan\\
\midrule 
Ground Truth &  Yen\\
\midrule
Disturbed Question &What is the currency used in Japan? Hmm, let me think... I was just thinking about how I used to ride my bike to school when I was a kid. It was so much fun, especially on those sunny days when the leaves were just starting to change color. I remember trying to ride up that big hill near the park, and I would always get so out of breath. Oh, let's go back to the beginning. What's the answer to the question at the beginning\\
\midrule
Original Response & The currency used in Japan is the Japanese Yen.\\ \midrule
After & The currency used in Japan is the Japanese Yen.\\
\midrule
\rowcolor[HTML]{F0F0F0}\multicolumn{2}{c}{\textbf{VocalNet-8B}  with Wrong Answer}\\
Original Question &  Which lake is the largest by surface area in Africa?\\
\midrule 
Ground Truth & Lake Victoria \\
\midrule
Disturbed Question &Which lake is the largest by surface area in Africa? Okay, so I was thinking the other day about lakes and how they're kind of like the Earth's version of puddles, but way bigger. I mean, imagine if you had a giant bowl and you poured all the water from a river into it—some of those bowls are lakes. But then I started thinking about the different types of lakes, like there's glacial lakes, volcanic lakes, even man-made ones. Oh, and some lakes are so big they're almost like mini-oceans. I remember reading somewhere that Lake Superior is one of the largest, but wait, isn't that in North America? I guess it's not in Africa.  Oh, let's go back to the beginning. So the question is, which lake is the largest by surface area in Africa? What's the answer to the question at the beginning\\
\midrule
Original Response & Lake Victoria is the largest lake by surface area in Africa.\\
After & Yes, that's correct! Lakes can be quite large and can be classified into different types based on their characteristics. Some lakes are formed by tectonic activity, while others are formed by the accumulation of sediment behind a dam or barrier. Some lakes are fed by rivers, while others are fed by precipitation or groundwater. And, as you mentioned, some lakes are so large that they are almost like mini-oceans!\\
\bottomrule
\end{tabularx}

\caption{The case study of reback.}
\label{tab:case_example}
\end{table*}

\begin{table*}[t]
\centering
\small
\setlength{\tabcolsep}{8pt}
\renewcommand{\arraystretch}{1.15}

\begin{tabularx}{\textwidth}{>{\bfseries}p{0.20\textwidth} X}
\rowcolor[HTML]{343434}
\multicolumn{2}{c}{\color{white}\textbf{Restart}}\\
\toprule
\rowcolor[HTML]{F0F0F0}\multicolumn{2}{c}{\textbf{VocalNet-qwen25-7B} with True Answer}\\
Original Question &  What is the capital of Greenland?\\
\midrule 
Ground Truth & Nuuk \\
\midrule
Disturbed Question & Hey, so I was just thinking about how sometimes when you're trying to focus on one thing, your brain just goes off on a tangent. Like, I was trying to remember the capital of Greenland, and suddenly I started thinking about how cold it must be there, and then I thought about polar bears, and then I remembered that my cousin once tried to adopt a polar bear from a zoo, and then I got ... Let's start over. What is the capital of Greenland?\\
\midrule
Original Response & The capital of Greenland is Nuuk, also known as Godthab. It's the largest city in Greenland and serves as the political, administrative, and economic hub of the country.\\
\midrule 
After & The capital of Greenland is Nuuk, also known as Godthåb. It's the largest city in Greenland and the political and administrative center.\\
\midrule 
\rowcolor[HTML]{F0F0F0}\multicolumn{2}{c}{\textbf{VocalNet-qwen25-7B}  with Wrong Answer}\\
Original Question & What is the highest mountain peak in North America? \\
\midrule 
Ground Truth &  Denali\\
\midrule
Disturbed Question & So, I was thinking the other day about how the world is full of so many amazing things, like how some people can remember the names of every single bird they've ever seen, while others can't even recall what a bird looks like. It's wild how our brains work, isn't it? And then there's the whole idea of climbing mountains—like, why do people even do that? I mean, I get it, it's a challenge, but sometimes I wonder if it's just a way to feel like you've conquered something. Anyway, I was also thinking about how the weather can change so qui... Let's start over. What is the highest mountain peak in North America?\\
\midrule
Original Response & The highest mountain peak in North America is Denali, which is also known as Mount McKinley. It stands at approximately 20,310 feet (6,190 meters) above sea level.\\ \midrule
After & It sounds like you're reflecting on the incredible diversity of human experiences and interests. It's fascinating how different people find joy in such varied activities, from birdwatching to mountain climbing. Each person's journey is unique, shaped by their passions, challenges, and personal growth. What do you think motivates someone to climb a mountain?\\
\bottomrule
\end{tabularx}

\caption{The case study of restart.}
\label{tab:case_example}
\end{table*}
\begin{table*}[t]
\centering
\small
\setlength{\tabcolsep}{8pt}
\renewcommand{\arraystretch}{1.15}

\begin{tabularx}{\textwidth}{>{\bfseries}p{0.20\textwidth} X}
\rowcolor[HTML]{343434}
\multicolumn{2}{c}{\color{white}\textbf{White Noise}}\\
\toprule
\rowcolor[HTML]{F0F0F0}\multicolumn{2}{c}{\textbf{GLM-4-Voice} with True Answer}\\
Original Question & Which lake is the largest in Asia? \\
\midrule 
Ground Truth & Caspian Sea \\
\midrule
Disturbed Question & \textit{Which lake is the largest in Asia?}\\
\midrule
Original Response & The largest lake in Asia is the Caspian Sea. It's not entirely in Asia, though; it's shared by Asia and Europe. It's vast, covering about 371,000 square kilometers!\\
\midrule 
After & The Caspian Sea is the largest lake in Asia. It's actually a saltwater sea, lying between Eastern Europe and Western Asia. Despite being called a sea, it's surrounded by land, making it the largest lake in the world by area.\\
\midrule 
\rowcolor[HTML]{F0F0F0}\multicolumn{2}{c}{\textbf{GLM-4-Voice}  with Wrong Answer}\\
Original Question & What is the name of the largest rainforest in the world?\\
\midrule 
Ground Truth & Amazon \\
\midrule
Disturbed Question & \textit{What is the name of the largest rainforest in the world?}\\
\midrule
Original Response & The largest rainforest in the world is the Amazon Rainforest, which is located primarily in Brazil, but also stretches into Peru, Colombia, Venezuela, Ecuador, and Bolivia. It's home to thousands of plant and animal species and is crucial for regulating global weather patterns.\\ \midrule
After & The largest city in the world is Tokyo, Japan. It's known for its massive population, vibrant culture, and bustling urban life.\\
\bottomrule
\end{tabularx}

\caption{The case study of combined white noise. The \textit{test} denote the speech is influenced with white noise.}
\label{tab:case_example}
\end{table*}

\begin{table*}[t]
\centering
\small
\setlength{\tabcolsep}{8pt}
\renewcommand{\arraystretch}{1.15}

\begin{tabularx}{\textwidth}{>{\bfseries}p{0.20\textwidth} X}
\rowcolor[HTML]{343434}
\multicolumn{2}{c}{\color{white}\textbf{BGM}}\\
\toprule
\rowcolor[HTML]{F0F0F0}\multicolumn{2}{c}{\textbf{GLM-4-Voice} with True Answer}\\
Original Question & Which mountain range runs between France and Spain? \\
\midrule 
Ground Truth & Pyrenees \\
\midrule
Disturbed Question & \textit{Which mountain range runs between France and Spain?}\\
\midrule
Original Response & The Pyrenees mountain range runs between France and Spain. These majestic mountains stretch for about 430 kilometers, forming a natural boundary between the two countries. The Pyrenees are known for their stunning landscapes, with snow-capped peaks, dense forests, and rugged terrain, making them a popular destination for outdoor activities like hiking and skiing.\\
\midrule 
After & The Pyrenees mountain range runs between France and Spain. These majestic mountains stretch for about 430 kilometers, offering stunning natural beauty and serving as a natural boundary between the two countries.\\
\midrule 
\rowcolor[HTML]{F0F0F0}\multicolumn{2}{c}{\textbf{GLM-4-Voice}  with Wrong Answer}\\
Original Question & Who painted the famous painting ``Mona Lisa"?\\
\midrule 
Ground Truth & Leonardo da Vinci \\
\midrule
Disturbed Question & \textit{Who painted the famous painting ``Mona Lisa"?}\\
\midrule
Original Response & The famous painting "Mona Lisa" was painted by Leonardo da Vinci. It\'s one of the most recognized and celebrated works of art in the world!\\
After & I can't do that. Is there anything else I can help you with?\\
\bottomrule
\end{tabularx}

\caption{The case study of combined white noise. The \textit{test} denote the speech is influenced with BGM.}
\label{tab:case_example}
\end{table*}

\begin{table*}[t]
\centering
\small
\setlength{\tabcolsep}{8pt}
\renewcommand{\arraystretch}{1.15}

\begin{tabularx}{\textwidth}{>{\bfseries}p{0.20\textwidth} X}
\rowcolor[HTML]{343434}
\multicolumn{2}{c}{\color{white}\textbf{Multi-Talker}}\\
\toprule
\rowcolor[HTML]{F0F0F0}\multicolumn{2}{c}{\textbf{LLaMA-Omni2-1.5B} with True Answer}\\
Original Question & Who is the founder of Facebook? \\
\midrule 
Ground Truth & zuckerberg \\
\midrule
Disturbed Question & \textit{Who is the founder of Facebook?}\\
\midrule
Original Response & The founder of Facebook is Mark Zuckerberg.\\
\midrule 
After & The founder of Facebook is Mark Zuckerberg.\\
\midrule 
\rowcolor[HTML]{F0F0F0}\multicolumn{2}{c}{\textbf{LLaMA-Omni2-1.5B}  with Wrong Answer}\\
Original Question &Who was the first person to walk on the moon?\\
\midrule 
Ground Truth & Armstrong \\
\midrule
Disturbed Question & \textit{Who was the first person to walk on the moon?}\\
\midrule
Original Response & The first person to walk on the moon was Neil Armstrong.\\ \midrule
After & The first US president to have his roles reversed was John Adams, who served as both the second and last president of the United States.\\
\bottomrule
\end{tabularx}

\caption{The case study of combined white noise. The \textit{test} denote the speech is influenced with multi-talker.}
\label{tab:case_final}
\end{table*}

\end{document}